\documentclass[final,5p,times,twocolumn]{elsarticle}

\usepackage{amsmath}
\usepackage{mathtools}
\usepackage{xcolor}
\usepackage{enumitem}
\usepackage{comment}
\usepackage{graphicx}
\usepackage{multirow}
\usepackage{array}
\usepackage{caption}
\usepackage{fontawesome5}
\usepackage{hyperref}
\usepackage{wrapfig}

\usepackage{amssymb}

\makeatletter
\def\ps@pprintTitle{%
    \let\@oddhead\@empty
    \let\@evenhead\@empty
    \def\@oddfoot{\hbox to \textwidth{\hfil\thepage\hfil}}%
    \let\@evenfoot\@oddfoot
}
\makeatother

\begin{document}

\begin{frontmatter}



\title{Explainable Artificial Intelligence: \\ A Survey of Needs, Techniques, Applications, and Future Direction}

\author[1]{Melkamu Mersha} 
\author[2]{Khang Lam}
\author[1]{Joseph Wood} 
\author[1]{Ali AlShami} 
\author[1]{Jugal Kalita}

\affiliation[1]{organization={College of Engineering and Applied Science, University of Colorado Colorado Springs},
            addressline={},
            postcode={80918},
            state={CO},
            country={USA}}

\affiliation[2]{organization={College of Information and Communication Technology, Can Tho University},
            addressline={}, 
            city={Can Tho},
            postcode={90000}, 
            country={Vietnam}}

\begin{abstract}
Artificial intelligence models encounter significant challenges due to their black-box nature, particularly in safety-critical domains such as healthcare, finance, and autonomous vehicles. Explainable Artificial Intelligence (XAI) addresses these challenges by providing explanations for how these models make decisions and predictions, ensuring transparency, accountability, and fairness. Existing studies have examined the fundamental concepts of XAI, its general principles, and the scope of XAI techniques. However, there remains a gap in the literature as there are no comprehensive reviews that delve into the detailed mathematical representations, design methodologies of XAI models, and other associated aspects. This paper provides a comprehensive literature review encompassing common terminologies and definitions, the need for XAI, beneficiaries of XAI, a taxonomy of XAI methods, and the application of XAI methods in different application areas. The survey is aimed at XAI researchers, XAI practitioners, AI model developers, and XAI beneficiaries who are interested in enhancing the trustworthiness, transparency, accountability, and fairness of their AI models.

\end{abstract}


\end{frontmatter}

\section{Introduction}
\label{sec:introduction}
Since the advent of digital computer systems, scientists have been exploring ways to automate human intelligence via computational representation and mathematical theory, eventually giving birth to a computational approach known as Artificial Intelligence (AI). AI and machine learning (ML) models are being widely adopted in various domains, such as web search engines, speech recognition, self-driving cars, strategy game-play, image analysis, medical procedures, and national defense, many of which require high levels of security, transparent decision-making, and a responsibility to protect information \cite{weller2019transparency,samek2017explainable}. Nevertheless, significant challenges remain in trusting the output of these complex ML algorithms and AI models because the detailed inner logic and system architectures are obfuscated by the user by design.

AI has shown itself to be an efficient and effective way to handle many tasks at which humans usually excel. In fact, it has become pervasive, yet hidden from the casual observer, in our day-to-day lives. As AI techniques proliferate, the implementations are starting to outperform even the best expectations across many domains \cite{shrivastava2023novelty}. Since AI solves difficult problems, the methodologies used have become increasingly complex. A common analogy is that of the black box, where the inputs are well-defined, as are the outputs. However, the process is not transparent and cannot be easily understood by humans. The AI system does not usually provide any information about how it arrives at the decisions it makes. The systems and processes used in decision-making are often abstruse and contain uncertainty in how they operate. Since these systems impact lives, it leads to an emerging need to understand how decisions are made. Lack of such understanding makes it difficult to adopt such a powerful tool in industries that require sensitivity or that are critical to the survival of the species.

\begin{figure}[h]
\centering
\includegraphics[width=\linewidth]{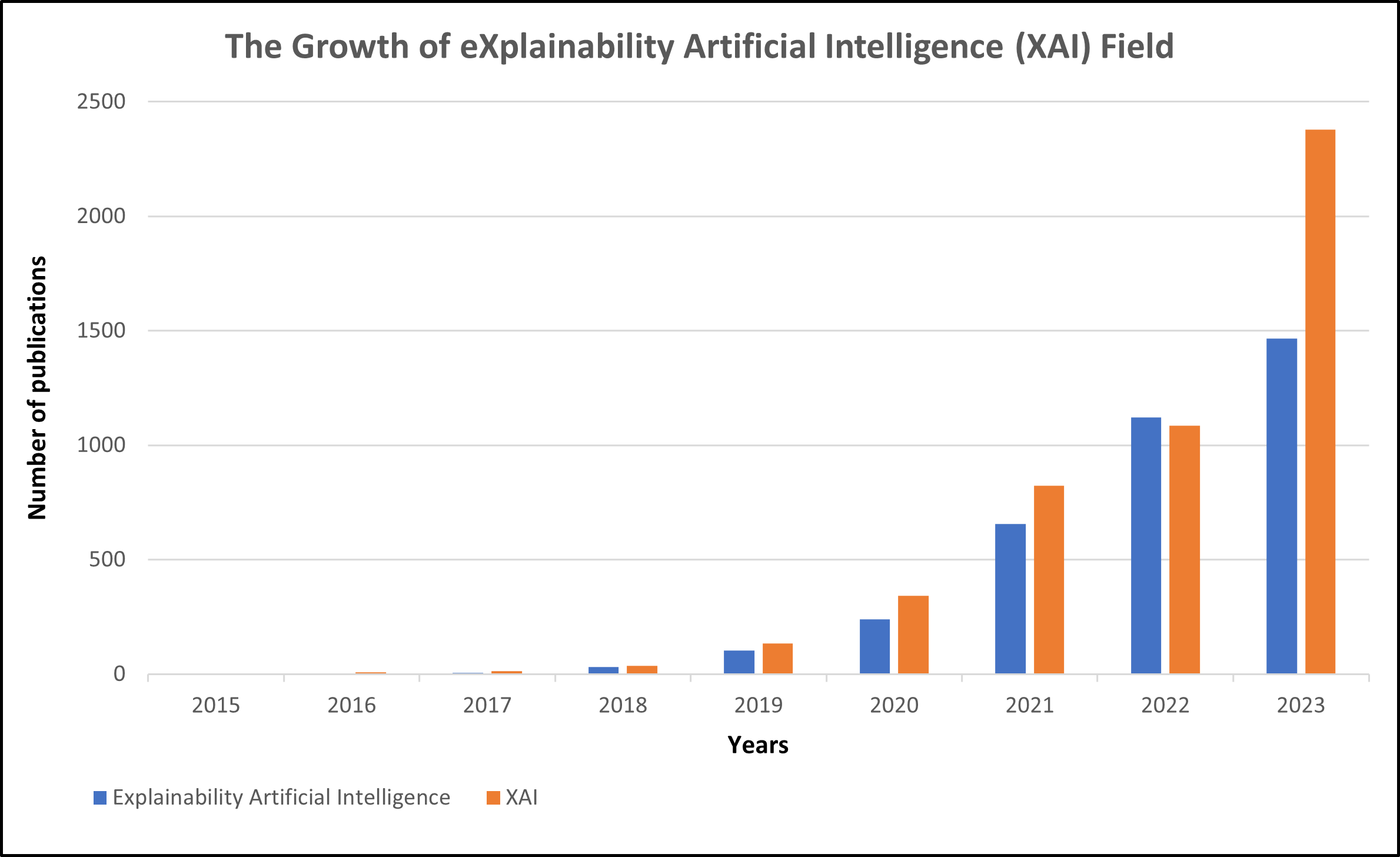}
\caption{Research publications in the explainability of artificial intelligence (XAI) field during the last few years}
\label{fig:myXAI}
\end{figure}

The black-box nature of AI models raises significant concerns, including the need for explainability, interpretability, accountability, and transparency. These aspects, along with legal, ethical, and safety considerations, are crucial for building trust in AI, not just among scientists but also among the wider public, regulators, and politicians who are increasingly attentive to new developments. With this in mind, there has been a shift from just relying on the power of AI to understanding and interpreting how AI has arrived at decisions, leading to terms such as transparency, explainability, interpretability, or, more generally, eXplainable Artificial Intelligence (XAI). A new approach is required to trust the AI and ML models, and though much has been accomplished in the last decades, the interpretability and black-box issues are still prevalent~\cite{vilone2020explainable, schwalbe2023comprehensive}. Attention given to XAI has grown steadily (Figure~\ref{fig:myXAI}), and XAI has attracted a thriving number of researchers, though there still exists a lack of consensus regarding symbology and terminology. Contributions rely heavily on their own terminology or theoretical framework~\cite{marcus2018deep}.

Researchers have been working to increase the interpretability of AI and ML models to gain better insight into black-box decision-making. Questions being explored include how to explain the decision-making process, approaches for interpretability and explainability, ethical implications, and detecting and addressing potential biases or errors \cite{weller2019transparency,guidotti2018survey}. These and other critical questions remain open and require further research. This survey attempts to address these questions and provide new insights to advance the adoption of explainable artificial intelligence among different stakeholders, including practitioners, educators, system designers, developers, and other beneficiaries. 

A significant number of comprehensive studies on XAI have been released. XAI survey publications usually focus on describing XAI basic terminology, outlining the explainability taxonomy, presenting XAI techniques, investigating XAI applications, analyzing XAI opportunities and challenges, and proposing future research directions. Depending on the goals of each study, the researchers may concentrate on specific aspects of XAI. Some outstanding survey papers and their main contributions are as follows. Gilpin et al.~\cite{gilpin2018explaining} defined and distinguished the key concepts of XAI, while Adadi and Berrada~\cite{adadi2018peeking} introduced criteria for developing XAI methods. Arrieta et al.~\cite{arrieta2020explainable} and Minh et al.~\cite{minh2022explainable} concentrated on XAI techniques. In addition to XAI techniques, Vilone and Longo~\cite{vilone2020explainable} also explored the evaluation methods for XAI. Stakeholders, who benefit from XAI, and their requirements were examined by Langer et al.~\cite{langer2021we}. Speith~\cite{speith2022review} performed studies on the common XAI taxonomies and identified new approaches to build new XAI taxonomies. R\"auker et al.~\cite{rauker2023toward} emphasized on inner interpretability of the deep learning models. The use of XAI to enhance machine learning models is investigated in the study of Weber et al.~\cite{weber2023beyond}. People have discussed the applications of XAI in a variety of domains and tasks~\cite{islam2022systematic} or specific domains, such as medicine~\cite{holzinger2019causability,lotsch2021explainable,gonzalez2023scoping}, healthcare~\cite{loh2022application,alam2023explainable,albahri2023systematic,saranya2023systematic}, and finance~\cite{saranya2023systematic}. Recently, Longo et al. \cite{longo2024explainable} proposed a manifesto to govern the XAI studies and introduce more than twenty open problems in XAI and their suggested solutions. 

\begin{figure}[h]
\centering
\includegraphics[width=\linewidth]{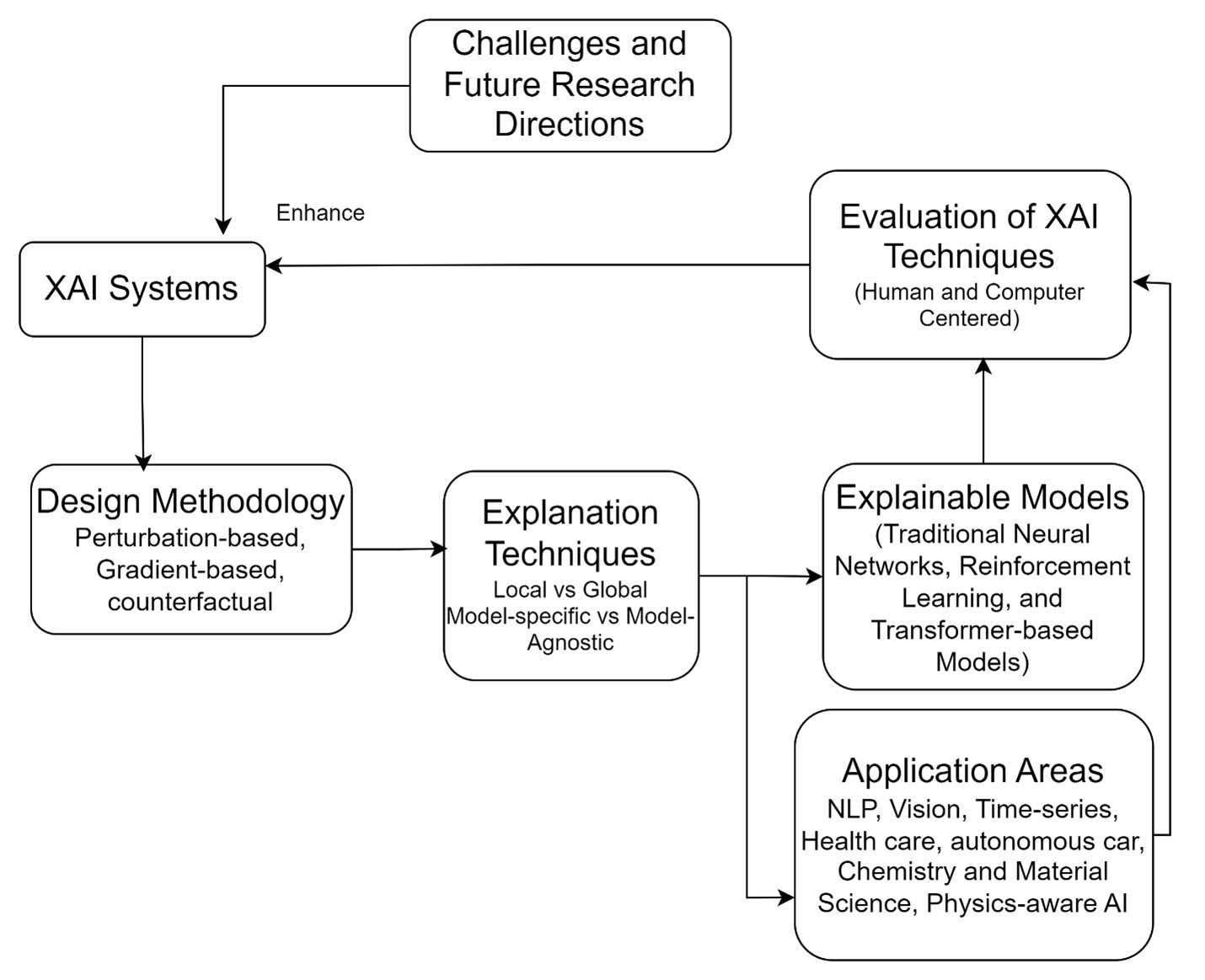}
\caption{`All you need here'-A comprehensive overview of XAI concepts.}
\label{fig:benfitsOFsurvey}
\end{figure}

Our systematic review carefully analyzes more than two hundred studies in the domain of XAI. This survey provides a complete picture of XAI techniques for beginners and advanced researchers. It also covered explainable models, application areas, evaluation of XAI techniques, challenges, and future directions in the domain, Figure~\ref{fig:benfitsOFsurvey}. The survey provides a comprehensive overview of XAI concepts, ranging from foundational principles to recent studies incorporating mathematical frameworks. In Figure  ~\ref{fig:benfitsOFsurvey}, our \textbf{``all you need here''} shows how our survey offers a clear and systematic approach, enabling readers to understand the multifaceted nature of XAI. To the best of our knowledge, this is the first work to comprehensively review explainability across traditional neural network models, reinforcement learning models, and Transformer-based models (including large language models and Vision Transformer models), covering various application areas, evaluation methods, XAI challenges, and future research directions.

The main contributions of our work are presented below:
\begin{itemize}
    \item Develop and present a comprehensive review of XAI that addresses and rectifies the limitations observed in previous review studies. 
    \item More than two hundred research articles were surveyed in this comprehensive study in the XAI field.
    \item Discuss the advantages and drawbacks of each XAI technique in depth.
    \item Highlight the research gaps and challenges of XAI to strengthen future works.
\end{itemize}

The paper is organized into eight sections: Section \ref{sectionBackground} introduces relevant terminology and reviews the background and motivation of XAI research. Section \ref{sectionTechniques} and Section \ref{sectionDiscussion} present types of explainability techniques and discussions on XAI techniques along different dimensions, respectively. Section \ref{sectionApplication} discusses XAI techniques in different applications. Section \ref{Evaluation} and \ref{FeatureReserechDirections} present XAI evaluation methods and future research direction, respectively. Section \ref{sectionConclusion} concludes the survey.

\section{Background and Motivation} \label{sectionBackground}
Black-box AI systems have become ubiquitous and are pervasively integrated into diverse areas. XAI has emerged as a necessity to establish trustworthy and transparent models, ensure governance and compliance, and evaluate and improve the decision-making process of AI systems. 

\subsection{Basic Terminology}
Before discussing XAI in-depth, we briefly present the basic terminology used in this work.

\textbf{AI systems} can perform tasks that normally require human intelligence \cite{bostrom2018ethics}. They can solve complex problems, learn from large amounts of data, make autonomous decisions, and understand and respond to challenging prompts using complex algorithms.

\textbf{XAI systems} refer to AI systems that are able to provide explanations for their decisions or predictions and give insight into their behaviors. In short, XAI attempts to understand ``WHY did the AI system do \textit{X}?''. This can help build	comprehensions about the influences on a model and	specifics about where a model succeeds or fails \cite{minh2022explainable}.

\textbf{Trust} is the degree to which people are willing to have confidence in the outputs and decisions provided by an AI system. A relevant question is: Does the user trust the output enough to perform real-world actions based upon it?~\cite{ribeiro2016should}.

\textbf{Machine learning} is a rapidly evolving field within computer science. It is a subset of AI that involves the creation of algorithms designed to emulate human intelligence by capturing data from surrounding environments and learning from such data using models, as discussed in the previous paragraph~\cite{el2015machine}. ML imitates the way humans learn, gradually improving accuracy over time based on experience. In essence, ML is about enabling computers to think and act with less human intervention by utilizing vast amounts of data to recognize patterns, make predictions, and take actions based on that data.

\textbf{Models and algorithms} are two different concepts. However, they are used together in the development of real-world AI systems. A model (in the context of machine learning) is a computational representation of a system whose primary purpose is to make empirical decisions and predictions based on the given input data (e.g., neural network, decision tree, or logistic regression). In contrast, an algorithm is a set of rules or instructions used to perform a task. The models can be simple or complex, and trained on the input data to improve their accuracy in decision-making or prediction. Algorithms can also be simple or complex, but they are used to perform a specific task without any training. Models and algorithms differ by output, function, design, and complexity \cite{moor1978three}.

\textbf{Deep learning} refers to ML approaches for building multi-layer (or ``deep'') artificial neural network models that solve challenging problems. Specifically, multiple (and usually complex) layers of neural networks are used to extract features from data, where the layers between the input and output layers are ``hidden'' and opaque \cite{saxe2021if}.

\textbf{A black-box model} refers to the lack of transparency and understanding of how an AI model works when making predictions or decisions. Extensive increases in the amount of data and performance of computational devices have driven AI models to become more complex, to the point that neural networks have arguably become as opaque as the human brain \cite{castelvecchi2016can}. The model accepts the input and gives the output or the prediction without any reasonable details about why and how the model made that prediction or decision. The black-box nature of AI models can be attributed to various factors, including model complexity, optimization algorithms, large and complex training data sets, and the algorithms and processes used to train the models. Deep neural AI models, in particular, exacerbate these concerns due to the design of deep neural networks (DNN), with components that remain hidden from human comprehension.

\subsection{Need for Explanation}
Black-box AI systems have become ubiquitous throughout society, extensively integrated in a diverse range of disciplines, and can be found permeating many aspects of daily activities. The need for explainability in real-world applications is multifaceted and essential for ensuring the performance and reliability of AI models while allowing users to work effectively with these models. XAI is becoming essential in building trustworthy, accountable, and transparent AI models to satisfy delicate application designs~\cite{doran2017does, angelov2021explainable}.

\textbf{Transparency}: Transparency is the capability of an AI system to provide understandable and reasonable explanations of a model’s decision or prediction process~\cite{vilone2020explainable, fan2021interpretability, dam2018explainable}. XAI systems explain how AI models arrive at their prediction or decision so that experts and model users can understand the logic behind the AI systems~\cite{holzinger2019causability, ali2023explainable}, which is crucial for trustworthiness and transparency. Transparency has a meaningful impact on people’s willingness to trust the AI system by using directly interpretable models or availing XAI system	explanations \cite{zhang2020effect}. 
For example, if on a mobile device, voice-to-text recognition systems produce wrong transcription, the consequences may not always be a big concern although it may be irritating. This may also be the case in a chat program like ChatGPT if the questions and answers are ``simple''. In this case, the need for explainability and transparency is less profound. In contrast, explainability and transparency are crucial in critical safety systems such as autonomous vehicles, medical diagnosis and treatment systems, air traffic	control systems, and military systems \cite{samek2017explainable}.

\textbf{Governance and compliance issues}: XAI enables governance in AI systems by confirming that decisions made by AI systems are ethical, accountable, transparent, and compliant with any laws and regulations. Organizations in domains such as healthcare and finance can be subject to strict regulations, requiring human understanding for certain types of decisions made by AI models \cite{weller2019transparency, jordan2015machine, zhang2021survey}. For example, if someone is denied a loan by the bank's AI system, he or she may have the right to know why the AI system made this decision. Similarly, if a class essay is graded by an AI and the student gets a bad grade, an explanation may be necessary. Bias is often present in the nature of ML algorithms' training process, which is sometimes difficult to notice. This raises concerns about an algorithm acting in a discriminatory way. XAI has been found to serve as a potential remedy for mitigating issues of discrimination in the realms of law and regulation \cite{doshi2017towards}. For instance, if AI systems use sensitive and protected attributes (e.g., religion, gender, sexual orientation, and race) and make biased decisions, XAI may help identify the root cause of the bias and give insight to rectify the wrong decision. Hence, XAI can help promote compliance with laws and regulations regarding data privacy and protection, discrimination, safety, and reliability.

\textbf{Model performance and debugging}:
XAI offers potential benefits in enhancing the performance of AI systems, particularly in terms of model design and debugging as well as decision-making processes \cite{samek2017explainable,zhang2018interpretable, samek2019explainable}. The use of XAI techniques facilitates the identification and selection of relevant features for developing accurate and practical models. These techniques help tune hyperparameters such as	choice of activation functions, number of layers, and learning rates to prevent underfitting or overfitting. The explanation also helps the developers with bias detection in the decision-making process. If the developers quickly detect the bias, they can adjust the system to ensure that outputs are unbiased and fair. XAI can enable developers to identify decision-making errors and correct them, helping develop more accurate and reliable models. Explanation can enable users to have more control over the models so as to be able to modify the input parameters and observe how parameter changes affect the prediction or decision. Users can also provide feedback to improve the model decision process based on the XAI explanation.

\textbf{Reliability}: ML models' predictions and outputs may result in unexpected failures. We need some control mechanisms or accountability to trust the	AI models' predictions and decisions. For example, a wrong decision by a medical or self-driving black-box may result in high risk for the impacted human beings \cite{doran2017does,zhang2021survey}.

\textbf{Safety}: In certain applications, such as self-driving cars or military drones, it is important to understand the decisions made by an AI system in order to ensure the safety, security, and the lives of humans involved \cite{amodei2016concrete}.

\textbf{Human-AI collaboration}: XAI can facilitate collaboration between humans and AI systems by enabling humans to understand the reasoning behind an AI's actions \cite{dietvorst2015algorithm}.

\subsection{Stakeholders of the XAI}
\begin{figure}[h]
\centering
\includegraphics[width=\linewidth]{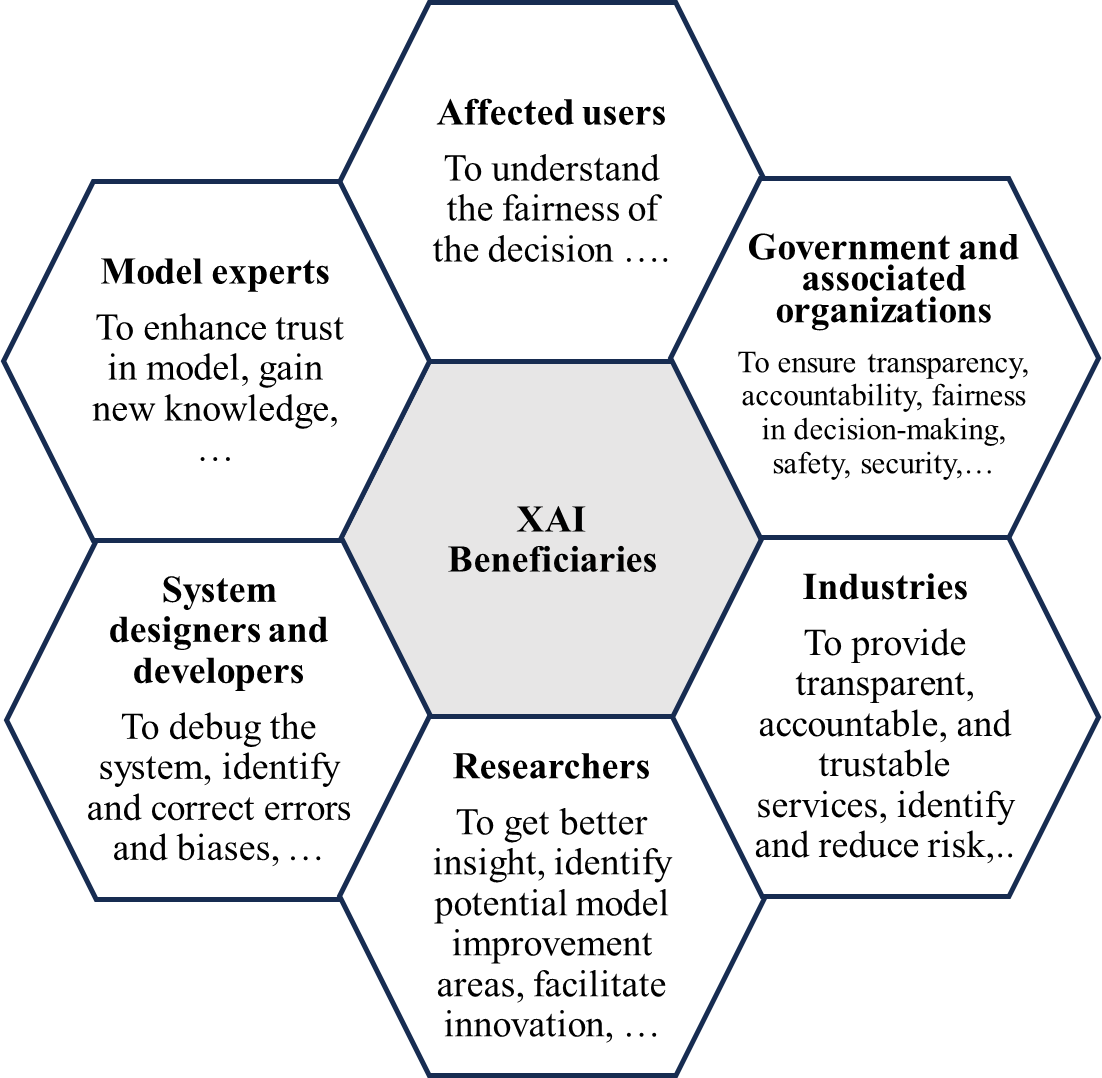}
\caption{XAI stakeholders/beneficiaries.}
\label{fig:XAIbeneficiaries}
\end{figure}

Broadly speaking, all users of XAI systems, whether direct or indirect, stand to benefit from AI technology. Some of the most common beneficiaries of the XAI system are identified in Figure~\ref{fig:XAIbeneficiaries}.

\textbf{Society:} XAI plays a significant role in fostering social collaboration and human-machine interactions \cite{samek2017explainable} by increasing the trustworthiness, reliability, and responsibility of AI systems, helping reduce the negative impacts such as unethical use of AI systems, discrimination, and biases. Hence, XAI promotes trust and the usage of models in society.

\textbf{Governments and associated organizations}: Governments and governmental organizations have become AI system users. Therefore, the government will be greatly benefited by XAI systems. XAI can help develop the government’s public policy decisions, such as public safety and resource allocation, by making them transparent, accountable, and explainable to society.

\textbf{Industries}: XAI is crucial for industries to provide	transparent, interpretable, accountable, and trustable services and decision-making processes. XAI can also help industries identify and reduce the risk of errors and biases, improve regulatory compliance, enhance customer trust and confidence, facilitate innovations, and increase accountability and transparency.

\textbf{Researchers and system developers}: The importance of XAI to researchers and AI system developers cannot be overstated, as it provides critical insights that lead to improved model performance. Specifically, XAI techniques enable them	to understand how AI models make decisions, and enable the identification of potential improvement and	optimization. XAI helps facilitate innovation and enhance the interpretability and explainability of the model. From a regulatory perspective, XAI can help enhance compliance with legal issues, in particular laws and regulations related to fairness, privacy, and security in the AI system. Finally, XAI can facilitate the debugging process critical to researchers and system developers, leading to the identification and correction of errors and biases.

\subsection{Interpretability vs. Explainability}
The concepts of \textit{interpretability} and \textit{explainability} are difficult to define rigorously. There is ongoing debate and research about the best ways to operationalize and measure these two concepts. Even terminology can vary or be used in contradictory ways, though the concepts of building comprehension about what influences a model, how influence occurs, and where the model performs well and fails, are consistent within the many definitions of these terms. Most studies at least agree that explainability and interpretability are related but distinct concepts. Previous work suggests that interpretability is not a monolithic concept but a combination of several distinct ideas that must be disentangled before any progress can be made toward a rigorous definition \cite{lipton2018mythos}. Explainability is seen as a subset of interpretability, which is the overarching concept that encompasses the idea of opening the black-box. 

The very first definition of interpretability in ML systems is ``the ability to explain or to present in understandable terms to a human'' \cite{doshi2017towards}, while explainability is ``the collection of features of the interpretable domain, that have contributed for a given example to produce a decision'' \cite{montavon2018methods}. As indicated by Fuhrman et al. \cite{fuhrman2022review}, interpretability refers to ``to understanding algorithm output for end‐user implementation'' and explainability refers to ``techniques applied by a developer to explain and improve the AI system''. Gurmessa and Jimma \cite{gurmessa2023comprehensive} defined these concepts as ``the extent to which human observers can understand the internal decision-making processes of the model'' and ``the provision of explanations for the actions or procedures undertaken by the model'', respectively.

According to Das et al., \cite{das2020opportunities}, interpretability and explainability are the ability ``to understand the decision-making process of an AI model'' and ``to explain the decision-making process of an AI model in terms understandable to the end user'', respectively. Another study defines these two concepts as ``the ability to determine cause and effect from a machine learning model'' and ``the knowledge to understand representations and how important they are to the model’s performance'', respectively \cite{gilpin2018explaining}. The AWS reports that interpretability is ``to understand exactly why and how the model is generating predictions'', whereas explainability is ``is how to take an ML model and explain the behavior in human terms''.

The goal of explainability and interpretability is to make it clear to a user how the model arrives at its output, so that the user can understand and trust the model’s decisions. However, there are no satisfactory functionally-grounded criteria or universally accepted benchmarks~\cite{marcinkevivcs2020interpretability}. The most common definitions of interpretable ML models are those that are easy to understand and describe, while explainable ML models can provide an explanation for their predictions or decisions~\cite{rudin2019stop}. A model that is highly interpretable is one that is simple and transparent, and whose behavior can be easily understood and explained by humans. Conversely, a model that is not interpretable is one that is complex and opaque, and whose behavior is difficult for humans to understand or explain~\cite{ribeiro2016model}.

In general, interpretability is concerned with how a model works, while explainability is concerned with why a model makes a particular prediction or decision. Interpretability is crucial because it allows people to understand how a model is making predictions, which can help build trust in the model and its results. Explainability is important because it allows people to understand the reasoning behind a model’s predictions, which can help identify any biases or errors in the model. 
Table~\ref{tabExampleXAItechniques} presents some representative XAI techniques and where they lie on the spectrum.

\begin{table*}[h]
  \begin{center}
    \caption{Examples of representative XAI techniques and where they lie on the spectrum.}        \label{tabExampleXAItechniques}
    \begin{tabular}{m{60pt}|m{70pt}|m{140pt}|m{175pt}} \hline
        \textbf{Spectrum} & \textbf{XAI techniques}&\textbf{How does it work}&\textbf{How to understand and explain} \\ \hline
		
        \multirow{7}{50pt}{Closer to Interpretability}&Linear regression&Use a linear relationship between the input features and the target variables to make	predictions.& Because the model is based on simple linear equations, it is easy for a human to understand and explain the relationship between the input features and the target variable. This is built into the model. \\ \cline{2-4}

        &Rule-based models& Use a set of explicit rules to make predictions.& Because the rules are explicit and transparent, these models are both interpretable and explainable, as it is easy for a human to understand and explain the rules that the model is using. \\ \hline		
        In the middle & Decision trees&Use a series of simple decision rules to make predictions. &These decision rules are based on the values of the input features. Because it is easy to trace	the model's predictions back to the input data and the decision rules, it's both interpretable and explainable.  \\ \hline
        
        \multirow{7}{50pt}{Closer to Explainability}&Feature importance analysis& Use an algorithm to identify the most important features in a model's prediction or	decision.& Because it provides a clear understanding of which features are most important, it is easy to trace	the model’s predictions back to the input features. This is usually post-hoc and not part of the model architecture, so more explainable then interpretable.  \\ \cline{2-4}
		
        &Local interpretable model-agnostic explanations& Use an approximate model to provide explanations for the predictions of a complex ML model. It works by approximating the complex model with a simple, interpretable model, and providing explanations based on the	simple model. &Because it provides explanations for the predictions of a complex model in a way that is	understandable to a human and is model-agnostic, LIME is explainable \\ \hline

	\end{tabular}
\end{center}  
\end{table*}

\section{Categories of Explainability Techniques} \label{sectionTechniques}
In this section, we introduce a taxonomy for XAI techniques and use specific criteria for general categorization. These explainability criteria, such as scope, stage, result, and function \cite{das2020opportunities, arrieta2020explainable, speith2022review}, are what we believe to be the most important because they provide a systematic and comprehensive framework for understanding and evaluating different XAI techniques. We have developed this taxonomy through rigorous study and analysis of existing taxonomies, along with an extensive review of research literature pertinent to explainable artificial intelligence. We categorize our reviewed papers by the scope of explainability and training level or stage. The explainability technique can be either global or local, and model-agnostic or model-specific, which can explain the model's output or function \cite{adadi2018peeking}.

\subsection{Local and Global Explanation Techniques}
Local and global approaches refer to the scope of explanations provided by an explainability technique. \textit{Local} explanations are focused on explaining predictions or decisions made by a specific instance or input to a model \cite{samek2017explainable, arrieta2020explainable}. This approach is particularly useful for examining the behavior of the model in relation to the local, individual predictions or decisions. 

\textit{Global} techniques provide either an overview or a complete description of the model, but such techniques usually require knowledge of input data, algorithm, and trained model~\cite{lipton2018mythos}. The global explanation technique needs to understand the whole structures, features, weights, and other parameters. In practice, global techniques are challenging to implement since complex models with multiple dimensions, millions of parameters, and weights are challenging to understand.

\subsection{Ante-hoc and Post-hoc Explanation Techniques}
Ante-hoc and post-hoc explanation techniques are two different ways to explain the inner workings of AI systems. The critical difference between them is the stage in which they are implemented \cite{guidotti2018survey}. The ante-hoc XAI techniques are employed during the training and development stages of an AI system to make the model more transparent and understandable, whereas the post-hoc explanation techniques are employed after the AI models have been trained and deployed to explain the model's prediction or decision-making process to the model users. Post-hoc explainability focuses on models which are not readily explainable by ante-hoc techniques. Ante-hoc and post-hoc explanation techniques can be employed in tandem to gain a more comprehensive comprehension of AI systems, as they are mutually reinforcing \cite{arrieta2020explainable}. Some examples of ante-hoc XAI techniques are decision trees, general additive models, and Bayesian models. Some examples of post-hoc XAI techniques are Local Interpretable Model-Agnostic Explanations (LIME)~\cite{ribeiro2016should} and Shapley Additive Explanations (SHAP)~\cite{lundberg2017unified}.

Arrieta et al.~\cite{arrieta2020explainable} classify the post-hoc explanation techniques into two categories:
\begin{itemize}
    \item \textbf{Model-specific approaches} provide explanations for the predictions or decisions made by a specific AI model, based on the model's internal working structure and design. These techniques may not apply to other models with varying architectures, since they are designed for specific models~\cite{vilone2020explainable}. However, a model-specific technique provides good insights into how the model works and makes a decision. For example, neural networks, random forests, and support vector machine models require model-specific explanation methods. The model-specific technique in neural networks provides more comprehensive insights into the network structure, including how weights are allocated to individual neurons and which neurons are explicitly activated for a given instance.
    \item \textbf{Model-agnostic approaches} are applied to all AI models and provide explanations of the models without depending on an understanding of the model's internal working structure or design. This approach is used to explain complex models that are difficult to explain using ante-hoc explanation techniques.	Model-agnostic approaches are model flexible, explanation flexible, and representation flexible, making them useful for a wide range of models. However, if the model is very complex, it may be hard to understand its behavior globally due to its flexibility and interpretability~\cite{ribeiro2016model}.
\end{itemize}

\subsection{Perturbation-based and Gradient-based XAI}
Perturbation-based and gradient-based methods are two of the most common algorithmic design methodologies for developing  XAI techniques. Perturbation-based methods operate by modifying the input data, while gradient-based methods calculate the gradients of the model's prediction with respect to its input data. Both techniques compute the importance of each input feature through different approaches and can be used for local and global explanations. Additionally, both techniques are generally model-agnostic.

Perturbation-based XAI methods use perturbations to determine the importance of each feature in the model's prediction process. These methods involve modifying the input data, such as removing certain input examples, masking specific input features, producing noise over the input features, observing how the model's output changes as a result, generating perturbations, and analyzing the extent to which the output is affected by the change of the input data. By comparing the original output with the output from the modified input, it is possible to infer which features of the input data are most important for the model's prediction~\cite{ribeiro2016should}. The importance of each feature value provides valuable insights into how the model made that prediction~\cite{lundberg2017unified}. Hence, the explanation of the model is generated iteratively using perturbation-based XAI techniques such as LIME, SHAP, and counterfactual.

\begin{figure*}[h]
\centering
\includegraphics[width=0.85\linewidth,height=0.35\textheight]{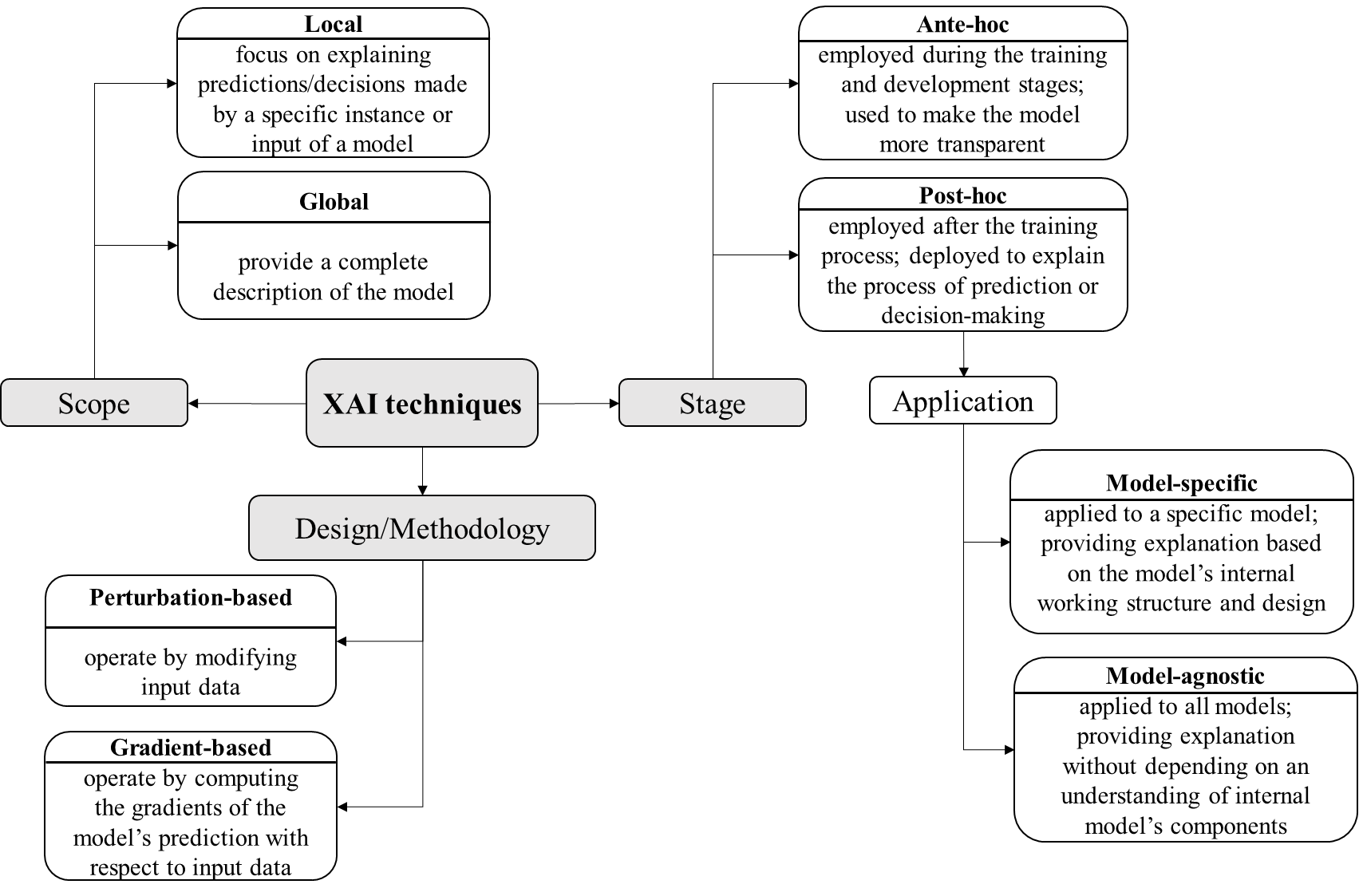}
\caption{Explainability taxonomy.} 
\label{fig:ExplainabilityTaxonomy}
\end{figure*}

Gradient-based XAI methods obtain the gradients of the model’s prediction with respect to its input features. These gradients reflect the sensitivity of the model's output to changes in each input feature \cite{ancona2017towards}. A higher gradient value for an input feature implies greater importance for the model’s prediction. Gradient-based XAI methods are valuable for their ability to handle high-dimensional input space and scalability for large datasets and models. These methods can help gain a deeper understanding of the model and detect errors and biases that decrease its reliability and accuracy, particularly in safety-critical applications such as health care and self-driving cars \cite{samek2017explainable}. Class activation maps, integrated gradients, and saliency maps are among the most commonly used gradient-based XAI methods. 
Figure~\ref{fig:ExplainabilityTaxonomy} presents a summary of explainability taxonomy discussed in this section.  

\begin{figure*}[h]
\centering
\includegraphics[width=0.8\linewidth,height=0.26\textheight]{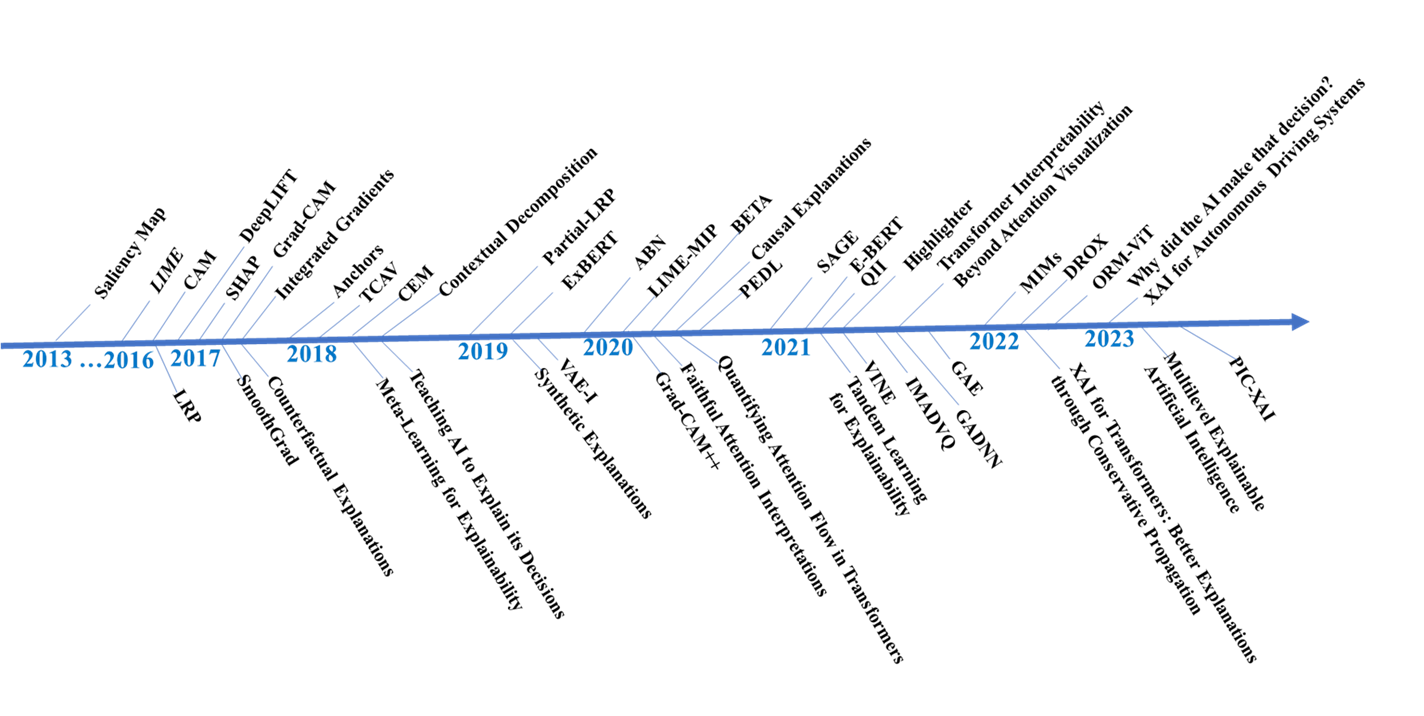}
 \caption{Chronological order of state-of-the-art XAI techniques.} 
\label{fig:StateofArts}
\end{figure*}

Figure~\ref{fig:StateofArts} illustrates a chronological overview of the state-of-the-art XAI techniques focused on in this survey. Perturbation-based methods like LIME, SHAP, Counterfactual explanations, and gradient approaches, including LRP, CAM, and Integrated Gradients, have been selected for detailed discussion in this context. They serve as foundational frameworks upon which other techniques are built, highlighting their significance in the field. ``Transformer Interpretability Beyond Attention Visualization'' \cite{chefer2021transformer} and ``XAI for Transformers: Better Explanations through Conservative Propagation'' \cite{ali2022xai} are foundational works for discussing transformer explainability, providing key insights and practices that serve as a baseline in this survey.

\section{Detailed Discussions on XAI Techniques} \label{sectionDiscussion}
XAI techniques differ in their underlying mathematical principles and assumptions, as well as in their applicability and limitations. We classify the widely used XAI techniques based on perturbation, gradient, and the use of the Transformer \cite{vaswani2017attention} architecture. The Transformer has become a dominant architecture in deep learning, whether it is in natural language processing, computer vision, time series data, or anything else. As a result, we include a separate section on Transformer explainability.

\subsection{Perturbation-based Techniques}
Perturbation-based XAI methods are used to provide local and global explanations of the black-box models by making small and controlled changes to the input data to gain insights into how the model made that decision. This section discusses the most predominant perturbation-based XAI techniques, such as LIME, SHAP, and Counterfactual Explanations (CFE), including their mathematical formulation and underlying assumptions. 

\subsubsection{LIME} \label{subsection:LIME}
A standard definition of a black-box model $f$, where the internal workings are unknown, is $f: X \rightarrow Y$, where $X$ and $Y$ represent the input and output spaces, respectively \cite{ das2020opportunities,ribeiro2018anchors}. Specifically, $x \in X$ denotes an input instance, and $y \in Y$ denotes the corresponding output or prediction. Let $X'$ be the set of perturbed and generated sample instances around the instance $x$ and $x' \in X'$, an instance from this set. Another function $g$ maps instances of $X'$ to a set of representations denoted as $Y'$ which are designed to be easily understandable or explainable: $g:X' \rightarrow Y'$, where $y' \in Y'$ is an output from the set of possible outputs in $Y'$. The use of interpretable instances in $Y'$ allows for clearer insights into the model's prediction processes.

LIME~\cite{ribeiro2016should} provides an explanation for each input instance $x$, where $f(x)=y' \approx y$ is the prediction of the black-box model. The LIME model is $g \in G$ where $g$ is explanation model that belongs to a set of interpretable models $G$. Let's say every $g \notin G$ is ``good enough'' to be interpretable. To prove this hypothesis, LIME uses three important arguments: a measure of complexity $\Omega(g)$ of the explanation, ensuring it remains simple enough for human understanding; a proximity measure $(\pi_{x}(z))$  that quantifies the closeness between the original instance $x$
and its perturbations; and a fidelity measure $\zeta (f,g,\pi_{x})$ which assesses how well $g$ approximates $ f$'s predictions, aiming for this value to be minimal to maximize the faithfulness of the explanation. The following formula achieves the explanation produced by LIME:
\begin{equation}
	\xi(x)=\text{argmin}~\zeta(f,g,\pi_x) + \Omega(g). \label{eq:LIME}
\end{equation}

\begin{figure}[h]
	\centering
	\includegraphics[width=\linewidth]{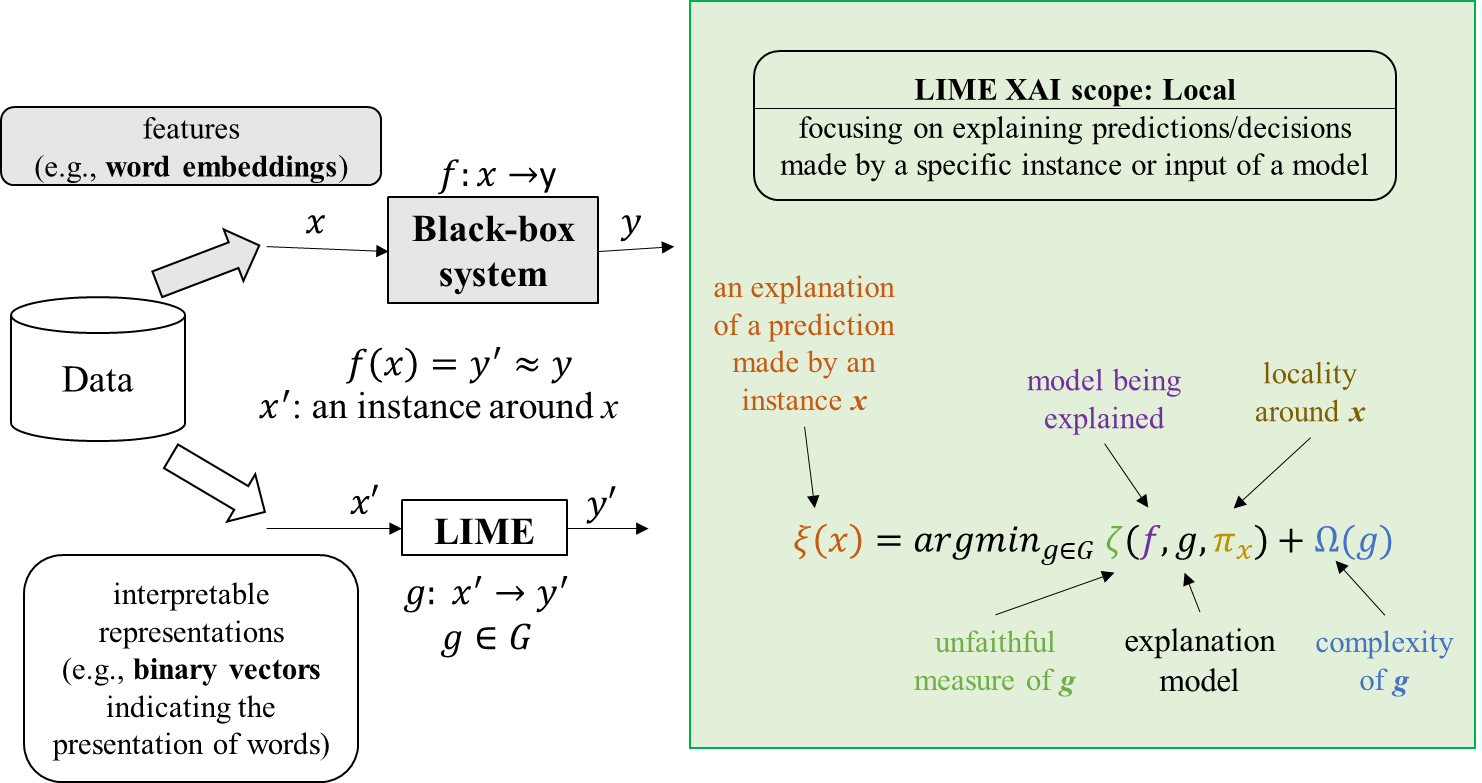}
	\caption{Illustration of LIME detailing instance-level interpretations for predictions by a black-box model.}
	\label{fig:LIME}
\end{figure}

Figure~\ref{fig:LIME} illustrates the LIME model for explaining a prediction of a black-box model based on an instance. LIME can be considered a model-agnostic technique for generating explanations
that can be used across different ML models. LIME is insightful in understanding the specific decisions of a model by providing local individual instance explanations, and in detecting and fixing biases by identifying the most influential feature for a particular decision made by a model~\cite{arrieta2020explainable}.

\subsubsection{SHAP}
SHAP~\cite{lundberg2017unified} is a model-agnostic method, applicable to any ML model, ranging from simple linear models to complex DNN. This XAI technique employs contribution values as means for explicating the extent to which features contribute to a model's output. The contribution value is then leveraged to explain the output of a given instance $x$. SHAP computes the average contribution of each feature through the subset of features by simulating the model's behavior for all combinations of feature values. The difference in output is computed when a feature is excluded or included in that output process. The subsequent contribution values give a measure of the feature relevance, which is significant to the model's output \cite{lundberg2017unified, ancona2019explaining}.

Assume $f$ is the original or black-box model, $g$ is the explanation model, $M$ is the number of simplified input features, $x$ is a single input, and $x'$ is a simplified input such that $x=h_x(x')$. Additive feature attribution methods, such as SHAP, have a linear function model explanation with binary variables. 
\begin{equation}
 g(x') = \phi_0 + \sum_{i=1}^{M} \phi_{i}z'_{i},  
\end{equation}
where $\phi_0$  is the default explanation when no binary features, $z'_i\in\{0,1\}^{M}$ and $\phi_i \in R$. The SHAP model explanation must satisfy three properties to provide high accuracy~\cite{lundberg2017unified}: (i) ``local accuracy'' requires that the explanation model $g(x')$ matches the original model $f(x)$, (ii) ``missingness'' which states, if the simplified inputs denote feature presence, then its attribute influence would be 0. More simply, if a feature is absent, it should have no impact on the model output, and (iii) ``consistency'' requires that if the contribution of a simplified input increases or stays the same (regardless of the other inputs), then the input's attribution should not decrease.

SHAP leverages contribution values to explain the importance of each feature to a model's output. The explanation is based on the Shapley values, which represent the average contribution of each feature over all possible subsets of features. 
However, in complex models, SHAP approximates the features of influences that may result in less accurate explanations. SHAP's explanation is model output dependent. If the model is biased, SHAP's explanation reflects the bias of the model behavior. 

\subsubsection{CFE}
CFE~\cite{wachter2017counterfactual} is used to explain the predictions made by the ML model using generated hypothetical scenarios to understand how the model's output is affected by changes in input data. The standard classification models are trained to find the the optimal set of weights $w$: 
\begin{equation}
    argmin_{\omega} \zeta(f_{\omega}(x_{i}), y_{i}) + \rho(w),  
\end{equation}
where $f$ is a model, $\rho$ is the regularizer to prevent overfitting in the training process, $y_{i}$ is the label for data point $x_{i}$, and $w$ represents the model parameters to be optimized. The argument $argmin$ is the value of the variable that minimizes the given function, then,
\begin{equation} 
argmin_{x'} max_{\lambda} \lambda(f_{\omega}(x') - y')^{2} + d(x_{i}, x').    
\end{equation}
This equation performs two different distance computations, comprising a quadratic distance between the model's output for the counterfactual $x'$ and the required output $y'$, and a distance $d$ between the input data $x_i$ to be explained and the counterfactual $x'$. The value of $\lambda$ can be maximized by solving for $x'$ iteratively and increasing the value of $\lambda$ until the closest solution is located. The distance function (viz., Manhattan distance)  $d$ should be carefully chosen based on task-specific requirements.

The CFE technique offers valuable insights into the decision-making process of a model, thereby aiding in the identification of biases and errors present in the black-box model. Importantly, the interpretation of the results and the insights gained from the counterfactual explanations can be model-agnostic, while the generation of the counterfactuals may not be.
In addition to being computationally expensive to generate counterfactuals, CFE is limited to individual instances and might not provide a general behavior of the model. CFE is data distribution-dependent, which may not be consistent if the training data is incomplete or biased. Finally, CFE is sensitive to ethical concerns if counterfactuals make suggestions (e.g., gender or race).

\subsection{Gradient-based Techniques}
Gradient-based techniques use the gradients of the output with respect to the input features.  
They can handle high-dimensional input space, are scalable for large datasets, provide deep insights into a model, and help detect errors and biases. Saliency Maps, Layer-wise Relevance BackPropagation (LRP), Class Activation Maps (CAM), and Integrated Gradients are the most common gradient-based XAI techniques and they are good frameworks for building other techniques.

\subsubsection{Saliency Map}
 
Simonyan et al. \cite{simonyan2013deep} utilized a saliency map for a model explanation for the first time in deep CNN. As a visualization technique, a saliency map, which is a model-agnostic technique, highlights important features in the image classification model by computing the output's gradients for the input image and visualizing the most significant regions of that image for the model's prediction.

Suppose an image $I_{0}$, a specific class $c$, and the CNN classification model with the class score function $S_c(I)$ (that is used to determine the score of the image) are analyzed. The pixels of $I_{0}$ are ranked based on their impact on this score $S_c(I_0)$. The linear score model for the class $c$ is obtained as follows:
\begin{equation}
S_c(I)=\omega_c^T + b_c, 
\end{equation}
where $I$ is a one-dimensional vectorized image, $\omega_c$ is the weight vector, and $b_c$ is the model's bias. The magnitude of weights $\omega$ specifies the relevance of the corresponding pixels of an image $I$ for class $c$. 
In the case of non-linear functions, $S_c(I)$ requires to be approximated based on the neighborhood of $I_0$ with the linear function using first-order Taylor expansion:
\begin{equation}
S_c(I) \approx \omega^T + b,    
\end{equation}
where $\omega$ is the derivative of $S_c$ with respect to the input image $I$ at the particular point in the image $I_0$:
\begin{equation}
\omega = \frac{\partial S_c}{\partial I}\bigg|_{I_{0}}.    
\end{equation}

The saliency map is sensitive to noise in the input data, which may lead to incorrect explanations. The saliency map is only applicable to gradient-based models. It provides a local explanation for individual predictions without suggesting the global behavior of a model. The explanation of saliency maps is sometimes ambiguous, where multiple features are highlighted, particularly in complex images~\cite{arrieta2020explainable}.

\subsubsection{LRP}
\begin{figure}[h]
	\centering
	\includegraphics[width=\linewidth]{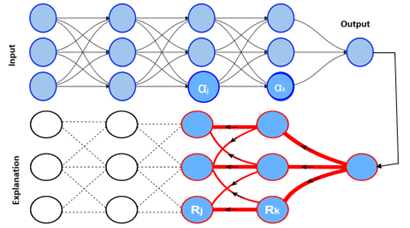}
	\caption{Visualizing the LRP technique (adapted from \cite{montavon2019layer}). Each neuron redistributes the relevance score to the lower layer ($R_j$) when it receives from, the higher layer ($R_k$).}
	\label{fig:LRP2}
\end{figure}

The main goal of LRP~\cite{bach2015pixel} is to explain each input feature's contribution to the model's output by assigning a relevance score to each neuron. LRP, as visualized in Figure~\ref{fig:LRP2}, propagates the relevance score backward through the network layers. It assigns a relevance score to each neuron, which allows for the determination of the contribution of each input feature to the output of the model.

LRP is subject to the conservation property, which means a neuron that receives the relevance score must be redistributed to the lower layer in an equal amount. Assume $j$ and $k$ are two consecutive layers, where layer $k$ is closer to the output layer. The neurons in layer $k$ have computed their relevance scores, denoted as  ($R_{k})_{k}$, propagating relevance scores to layer $j$. Then, propagated relevance score to neuron $R_{j}$ is computed using the following formula~\cite{montavon2019layer}:
\begin{equation}
R_j = \sum_k \frac{z_{jk}}{\sum_jz_{jk}} R_k,
\end{equation}
where $z_{jk}$ is the contribution of neuron $j$ to $R_k$ and $\sum_jz_{jk}$ is used to enforce the conservation property. In this context, a pertinent question arises as to how do we determine $z_{jk}$, which represents the contribution of a neuron $j$ to a neuron $k$ in the network, is ascertained? LRP uses three significant rules to address this question \cite{bach2015pixel}. 
\begin{itemize}
    \item The basic rule redistributes the relevance score to the input features in proportion to their positive contribution to the output.
    \item The Epsilon rule uses an $\epsilon$ to diminish relevance scores when contributions to neuron $k$ are weak and contradictory. 
    \item The Gamma rule uses a large value of $\gamma$ to reduces negative contribution or to lower noise and enhance stability.
\end{itemize}

Overall, LRP is faithful, meaning that it does not introduce any bias into the explanation \cite{bach2015pixel}. This is important for ensuring that the explanations are accurate and trustworthy. LRP is complex to implement and interpret, which requires a good understanding of the neural networks' architecture. It is computationally expensive for large and complex models to compute the backpropagating relevance scores throughout all layers of the networks. LRP is only applicable to backpropagation-based models like neural networks. It requires access to the internal structure and parameters of the model, which is sometimes impossible if a model is proprietary. LRP is a framework for other XAI techniques. However, there is a lack of standardization, which leads to inconsistent explanations through different implementations. 

\subsubsection{CAM}
\begin{figure}[!h]
\centering
\includegraphics[width=\linewidth]{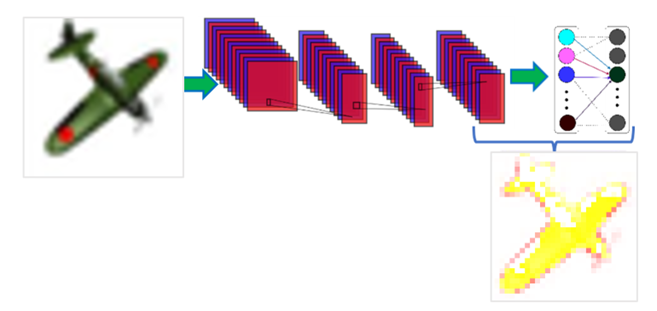}
\caption{The predicted class score is mapped back to the previous convolutional layer to generate the class activation maps (input image from CIFAR 10 dataset).}
\label{fig:CAM}
\end{figure}

CAM~\cite{zhou2016learning} is an explanation technique typically used for CNN and deep learning models applied to image data.  
For example, CAM can explain the predictions of a CNN model by indicating which regions of the input image the model is focusing on, or it can simply provide a heatmap for the output of the convolutional layer, as shown in Figure~\ref{fig:CAM}.

Let $f_k(x, y)$ denote the activation of unit $k$ in the last convolutional layer at location $(x, y)$ in the given image, the global average pooling is computed by $\sum_{x,y}f_k(x, y)$. Then, the input to the softmax function, called $S_c$, for a given class $c$ is obtained using the following formula:
\begin{equation}
    S_c=\sum_k w_k^c \sum_{x,y}f_k(x, y) =\sum_{x,y}\sum_k w_k^c f_k(x, y),
\end{equation}
where $w_k^c$ is the weight relating to the class $c$ for unit $k$. 
We can compute the class activation map $M_c$ for class $c$ of each special element as $\sum_{k}w_{k}^{c} f_{k}(x, y)$. 
Therefore, $S_{c}$ for a given class $c$ can be rewritten: 
\begin{equation}
   S_{c}=\sum_{x,y}M_{c}(x, y).
\end{equation}
In the previous formula, $M_{c}(x, y)$ shows the significance of the activation at spatial location $(x, y)$, and it is critical to determine the class of the image to class $c$.

CAM is a valuable explanation technique for understanding the decision-making process of deep learning models applied to image data. However, it is important to note that CAM is model-specific, as it requires access to the architecture and weights of the CNN model being used. 

\subsubsection{Integrated Gradients}
 Integrated Gradients \cite{sundararajan2017axiomatic} provides insights into the input-output behavior of DNNs which is critical in improving and building transparent ML models. Sundararajan et al. \cite{sundararajan2017axiomatic} strongly advocated that all attribution methods must adhere to two axioms. 
The \textit{Sensitivity} axiom is defined such that ``an attribution method satisfies Sensitivity if for every input and baseline that differ in one feature but have different predictions, then the differing feature should be given a non-zero attribution''. The violation of the Sensitivity axiom may expose the model to gradients being computed using non-relevant features. Thus, it is critical to control this sensitivity violation to assure the attribution method is in compliance. The \textit{Implementation Invariance} axiom is defined as ``two networks are functionally equivalent if their outputs are equal for all inputs, despite having very different implementations. Attribution methods should satisfy Implementation Invariance, i.e., the attributions are always identical for two functionally equivalent networks''. Suppose two neural networks perform the same task and generate identical predictions for all inputs. Then, any attribution method used on them should provide the same attribution values for each input to both networks, regardless of the differences in their implementation details. This ensures that the attributions are not affected by small changes in implementation details or architecture, thus controlling inconsistent or unreliable outputs. In this way, Implementation Invariance is critical to ensuring consistency and trustworthiness of attribution methods.

Consider a function F: $R^{n} \rightarrow$ [0, 1], which represents a DNN. We take $x \in R^{n}$ to be the input instance and $x' \in R^{n}$ be the baseline input. In order to produce a counterfactual explanation, it is important to define the baseline as the absence of a feature in the given input. However, it may be challenging to identify the baseline in a very complex model. For instance, the baseline for image data could be black images, while for NLP data, it could be a zero embedding vector, which is a vector of all zeroes used as a default value for words not found in the vocabulary. To obtain Integrated Gradients, we consider the straight-line path (in $R^{n}$) from $x'$ (the baseline) to the input instance $x$ and compute the gradients at all points along this path. The collection of these gradients provides the Integrated Gradients. In other words, Integrated Gradients can be defined as the path integral of the gradients along the straight-line path from $x'$ to the input instance $x$.

The gradient of F($x$) along the $i^{th}$ dimension is given by $\frac{\partial F(x)} { \partial x_{i}}$, 
leading to the Integrated Gradient ($IG$) along the $i^{th}$ dimension for an input $x$ and baseline $x'$ to be described as:
\begin{equation}
IG_{i}(x) =(x_{i}-x'_{i})\times \int^{1}_{\alpha=0}\frac{\partial F(x' + \alpha \times (x-x'))}{\partial x_{i}} d\alpha. \label{eq:IG02}
\end{equation}

The Integral Gradient XAI method satisfies several important properties, such as sensitivity, completeness, and implementation details. It can be applied to any differential model, making it a powerful and model-specific tool for explanation in a DNN \cite{sundararajan2017axiomatic}.

\subsection{XAI for Transformers}
Transformers \cite{vaswani2017attention} have emerged as the dominant architecture in Natural Language Processing (NLP), computer vision, multi-modal reasoning tasks, and a diverse and wide array of applications such as visual question answering, image-text retrieval, cross-modal generation, and visual commonsense reasoning~\cite{chefer2021generic}. Predictions and decisions in Transformer-based architectures heavily rely on various intricate attention mechanisms, including self-attention, multi-head attention, and co-attention. 
Explaining these mechanisms presents a significant challenge due to their complexity. In this section, we explore the interpretability aspects of widely adopted Transformer-based architectures. 

Gradient$\times$Input~\cite{ancona2019explaining}, \cite{shrikumar2017learning} and LRP \cite{bach2015pixel} XAI techniques have been extended to explain Transformers \cite{voita2019analyzing}, \cite{wu2021explaining}. Attention rollouts \cite{abnar2020quantifying} and generic attention are new techniques to aggregate attention information \cite{chefer2021generic} to explain the Transformers. LRP \cite{bach2015pixel}, Gradient$\times$Input \cite{ancona2017towards}, Integrated Gradients~\cite{sundararajan2017axiomatic}, and SHAP \cite{lundberg2017unified} are designed based on the conservation axiom for the attribution of each feature. The conservation axiom states that each input feature contributes a portion of the predicted score at the output. LRP is employed to assign the model's output back to the input features by propagating relevance scores backward through the layers of a neural network, measuring their corresponding contributions to the final output.  The relevance scores signify how much each feature at each layer contributes to the final prediction and decision.

The LRP framework is a baseline for developing various relevance propagation rules. Let's start the discussion by embedding Gradient$\times$Input into the LRP framework to explain Transformers \cite{ali2022xai}. Assume $(x_{i})_{i}$ and $(y_{j})_{j}$ represent the input and output vectors of the neurons, respectively,  and $f$ is the output of the model. Gradient$\times $Input attributions on these vector representations can be computed as:
\begin{equation}
  R(x_{i})= x_{i} \cdot (\partial f / \partial x_{i}) \text{ and } R(y_{j})= y_{j} \cdot (\partial f / \partial y_{j}). \label{eq:Transformer1} 
\end{equation}
The gradients at different layers are computed using the chain rule. This principle states that the gradient of the function $f$ with respect to an input neuron $x_{i}$ can be expressed as the sum of the products of two terms: the gradients of all connected neurons  $y_{j}$ with respect to $x_{i}$ and the gradients of the function $f$ with respect to those neurons $y_{j}$. This is mathematically represented as follows:

\begin{equation}
   \frac{\partial f} { \partial x_{i}} = \sum_{j}\frac{\partial f }  {\partial y_{j}} \frac{\partial y_{j} }  {\partial x_{i}}  . \label{eq:Transformer3} 
\end{equation}
We can convert the gradient propagation rule into an equivalent relevance propagation by inserting equation (\ref{eq:Transformer1}) into equation (\ref{eq:Transformer3}):
\begin{equation}
    R(x_{i}) = \sum_{j} \frac{\partial y_{j} }  {\partial x_{i}}  \frac{x_{i}} {y_{j}} R(y_{j}), \label{eq:Transformer4} 
\end{equation}
with the convention 0/0=0. We can prove that \text{$\sum_{i} R(x_{i})=\sum_{j} R(y_{j})$} easily, and if this condition holds true, conservation also holds true. However, Transformers break this conservation rule. The following subsections discuss methods to improve propagation rule \cite{ali2022xai}.

\subsubsection{Propagation in Attention Heads}
Transformers work based on Query ($Q$), Key ($K$), and Value~($V$) matrices, and consider the attention head, which uses these core components \cite{vaswani2017attention}. The attention heads have the following structure:
\begin{equation}
Y=softmax(\frac{1}{\sqrt{d_{k}}}(X'W_{Q})(XW_{K})^\tau )X, \label{eq:Transformer5}    
\end{equation}
where $X=(x_{i})_{i}$ and $X'=(x'_{j})_{j}$ are input sequences,  \text{$Y=(y_{j})_{j}$} is the sequence of the output, $W_{\{Q,K,V\}}$ are learned projection matrices, and $d_{k}$ is the dimensionality of the Key-vector. The previous equation is rewritten as follows:
\begin{equation}
y_{j}=\sum_{i} x_{i} p_{ij},    
\end{equation}
where $y_{j}$ is the output, $p_{ij}=\frac{exp(q_{ij})}{\sum_{i'}exp(q_{i'j})}$ is the softmax computation,
and $q_{ij}=\frac{1}{\sqrt{d_{k}}}x_{i}^\tau W_{K}W_{Q}^\tau x'_{j}$ is the matching function between the two input sequences.

\subsubsection{Propagation in LayerNorm}
LayerNorm or Layer normalization is the crucial component in Transformers used to stabilize and improve the training of models. LayerNorm is involved in the centering and standardization of key operations, defined as follows:
\begin{equation}
y_{i}= \frac{x_{i}-E[x]}{\sqrt{\epsilon + Var[x]}},     
\end{equation}
where $E[\cdot]$ and $Var[\cdot]$ represent the mean and variance overall activation of the corresponding channel. The relevance propagation associated with Gradient$\times$Input is represented by the conservation equation:
\begin{equation}
\sum_{i}R(x_{i})=(1-\frac{Var[x]}{\epsilon+Var[x]}) \sum_{i}R(y_{i}),    
\end{equation}
where $R(y_{j})=x_{j}^\tau (\partial f/\partial y_{j})$.
The implied propagation rules in attention heads and LayerNorm in equation (\ref{eq:Transformer4}) are replaced by ad-hoc propagation rules to ensure conservation. Hence, we make a locally linear expansion of \textit{attention head} by observing the gating terms $p_{ij}$ as constant, and these terms are considered as the weights of a linear layer which is locally mapping the input sequence $x$ into the output sequence $y$. As a result, we can use the canonical LRP rule for linear layers as follows:
\begin{equation}
R(x_{i})=\sum_{j}\frac{x_{i}p_{ij}}{\sum_{i'}x_{i'}p_{i'j}}R(y_{j}). 
\end{equation}

Recent studies such as  Attention rollouts \cite{abnar2020quantifying}, generic attention \cite{chefer2021generic}, and Better Explanations through Conservative Propagation \cite{ali2022xai} have provided empirical evidence that it is possible to improve the explainability of Transformers.

\subsection{Explainability in Reinforcement Learning}

Reinforcement Learning (RL) involves applications across various domains, including safety-critical areas like autonomous vehicles, healthcare, and energy systems \cite{rana2023safety, yu2021reinforcement}. In the domain of autonomous vehicles, RL is employed to refine adaptive cruise control and lane-keeping features by learning optimal decision-making strategies from simulations of diverse traffic scenarios \cite{ye2019automated}. Explainability in reinforcement learning concerns the ability to understand and explain the rationale behind the decisions made and actions taken by reinforcement learning models within their specified environments \cite{vouros2022explainable, madumal2020explainable, puiutta2020explainable}. Post-hoc explanations, such as SHAP and LIME, can help us understand and explain which features are most important for the decision-making process of an RL agent \cite{heuillet2022collective, heuillet2021explainability}. Example-based explanation methods, like trajectory analysis, help us to get insights into the decision-making process of the RL model by examining specific trajectories, such as sequences of states, actions, and rewards \cite{zhang2023learning}. Visualization techniques enable us to understand and interpret the RL models by visually representing the model’s decision-making processing \cite{wells2021explainable}. Several explainability methods exist to interpret reinforcement learning models, including saliency maps, counterfactual explanations, policy distillation, attention mechanisms, 	Human-in-the-loop, query system, and natural language explanations \cite{alharin2020reinforcement}. Explainability in reinforcement learning is crucial, particularly for safety-critical domains, due to the need for trust, safety assurance, regulatory compliance, ethical decision-making, model debugging, collaborative human-AI interaction, accountability, and AI model adoption and acceptance \cite{vouros2022explainable,chamola2023review, lai2021towards}.

\subsection{Summary}
Applying XAI techniques can enhance transparency and trust in AI models by explaining their decision-making and prediction processes. These techniques can be classified into categories such as local or global, post-hoc or ante-hoc, model-specific or model-agnostic, and perturbation or gradient methodology. We have added a special subsection for reinforcement learning and Transformers due to their popularity and profound impact on applications of deep learning in a wide variety of areas. Table~\ref{tabSummaryXAItechniques} summarizes the reviewed XAI techniques discussed. 
\begin{table*}[!h]
  \begin{center}
    \caption{The XAI techniques discussed, the methods used, their advantages and disadvantages.}        \label{tabSummaryXAItechniques}
    \begin{tabular}{m{40pt}|m{25pt}|m{45pt}|m{110pt}|m{70pt}|m{150pt}}\hline
        \textbf{Technique} & \textbf{Scope}&\textbf{Application}&\textbf{Method used}& \textbf{Advantages} & \textbf{Disadvantages} \\ \hline        
        \multicolumn{6}{c}{\textbf{Perturbation-based Techniques}}\\ \hline    
        LIME~\cite{ribeiro2016should} & Local &Model-Agnostic&-Explain a prediction of a black-box model based on an instance. &-Understanding the specific decisions of a model, \newline -Detecting and fixing biases.& -Computationally expensive, \newline -Less effective for complex and high-dimensional data. \\ \hline
   
        SHAP~\cite{lundberg2017unified} &Local &Model-Agnostic&-Leverage contribution values to explain the importance of each feature to a model's output.&-Applicable to any ML model.&-Computationally expensive, \newline -Need to be more scalable for highly complex models and large data, \newline
        Less accurate explanation in complex models. \\  \hline
        CFE~\cite{wachter2017counterfactual} &Local/ Global &Model-Agnostic&-Use generated hypothetical scenarios to understand how the model's output is affected by changes in input data.&-Offer valuable insights into the decision-making process of a model.& -Computationally expensive to generate counterfactuals, \newline   -Limited to individual instances and does not dicuss a general behavior of the model, \newline -Not consistent if the training data are incomplete or biased, \newline   -Sensitive to ethical concerns if counterfactuals make suggestions. \\  \hline
        \multicolumn{6}{c}{\textbf{Gradient-based methodology}}\\ \hline
        Saliency Map~\cite{simonyan2013deep} &Local/ Global &Model-Agnostic&-Highlight important features by obtaining output's gradients for input image, \newline -Visualize the most significant regions of that image for the model's prediction. &-Quick insights \newline -Applicable to various models&-Sensitive to noise in the input data, \newline-Only applicable to gradient-based models, \newline-Sometimes ambiguous, where multiple features are highlighted, particularly in complex images. \\ \hline 
        LRP~\cite{montavon2019layer} &Local/ Global &Model-specific&-Explain each input feature's contribution to the model's output by assigning a relevance score to each neuron.&-Faithful, \newline -Does not introduce any bias into the explanation.&-Complex to implement and interpret, \newline -Computationally expensive for large and complex models, \newline -Only applicable to backpropagation-based models, \newline -Lack of standardization, \newline -Inconsistent explanations through different implementations. \\ \hline
        CAM~\cite{zhou2016learning} &Local &Model-Specific&-Explain predictions of CNN by indicating which regions of the input image the model is focusing on, or it can simply provide a heatmap for the output of the convolutional layer.& -A valuable XAI technique for understanding the decision-making process of DL models applied to image data.&-Requires access to the architecture and weights of the model used. \\  \hline  
        Integrated Gradients ~\cite{sundararajan2017axiomatic} &Local & Model-Specific&-Attribution methods must adhere to Sensitivity and Implementation Invariance axioms. &-Satisfies several important properties,\newline-Applies to any models. &-Sensitivity to initialization \newline -Computational intensity \\  \hline   
       
        LRP- Conservation~\cite{ali2022xai} & Local & Model-Specific & -Explain predictions of the Transformers & -Granular insights \newline -Clear attribution & -Complexity and computational cost\\  \hline
        
        \end{tabular}
    \end{center}  
\end{table*}

\section{XAI Techniques in Application Areas} \label{sectionApplication}

The area of XAI has been gaining attention in recent years due to the growing need for transparency and trust in ML models \cite{ribeiro2016should, lundberg2017unified}. XAI techniques are being used to explain the predictions of ML models \cite{doshi2017towards,samek2017explainable, gilpin2018explaining}. These techniques can help identify errors and biases that decrease the reliability and accuracy of the models. This section explores the different XAI techniques used in natural language processing, computer vision, and time series analysis, and how they contribute to improving the trust, transparency, and accuracy of ML models in different application areas.

\subsection{Explainability in Natural Language Processing}
Natural language processing employs ML, as it can help efficiently handle, process, and analyze vast amounts of text data generated daily through areas such as human-to-human communication, chatbots, emails, and context generation software, to name a few \cite{torfi2020natural}. One barrier to implementation is that such data are usually not inherently clean, and preprocessing and training are essential tasks for achieving accurate results with language models \cite{jurafskyspeech}.  In NLP, language models can be classified into three categories: transparent architectures, neural network (non-Transformer) architectures, and transformer architectures.  Transparent models are straightforward and easy to understand due to their clear processing paths and direct interpretability. Models based on neural network (non-Transformer) architectures are often termed ``black boxes'' due to their multi-layered structures and non-linear processing. Transformer architectures utilize self-attention mechanisms to process sequences of data. The increased complexity and larger number of parameters often make transformer-based models less interpretable, requiring advanced techniques to explain their decision-making processes.   

\begin{figure}[h]
\centering
\includegraphics[width=\linewidth]{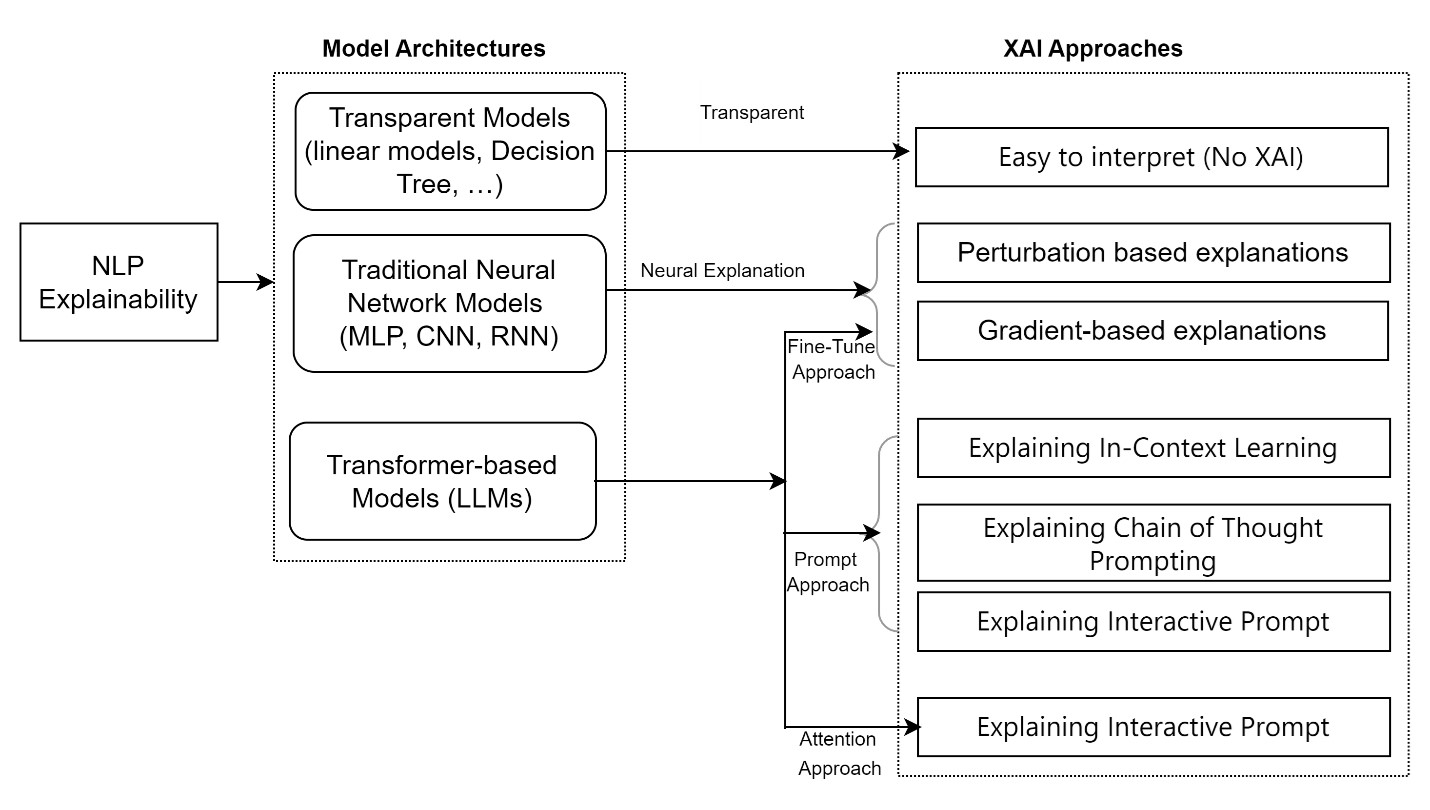}
\caption{Taxonomy of explainability in natural language processing.}
\label{fig:NLP_taxonomy}
\end{figure}

The success of XAI techniques used in NLP applications is heavily dependent on the quality of preprocessing and the type of text data used \cite{samek2017explainable, usuga2022using}. This is important because XAI is critical to developing reliable and transparent NLP models that can be employed for real-world applications by allowing us to understand how a model arrived at a particular decision. This section reviews some of the most common XAI techniques for NLP. Figure~\ref{fig:NLP_taxonomy} presents a taxonomy of NLP explainability.

\subsubsection{Explaining Neural Networks and Fine-Tuned Transformer Models: Insights and Techniques}
Transparent models are easy to interpret because their internal mechanisms and decision-making processes are designed to be inherently understandable \cite{arrieta2020explainable}. Perturbation-based and gradient-based techniques are the most commonly employed approaches for explaining neural network-based models and fine-tuned transformer-based models. In this subsection, we discuss some of the most common XAI techniques for neural network-based models and fine-tune Transformer-based models, used in NLP.

\begin{figure}[h]
\centering
\includegraphics[width=\linewidth]{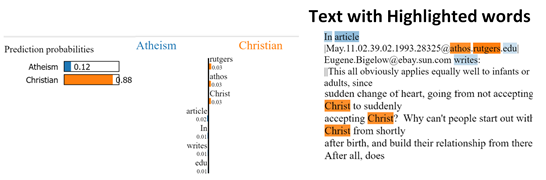}
\caption{LIME feature importance scores visualization.}
\label{fig:LIME_NLP}
\end{figure}

\textbf{\textit{LIME: }} 
Discussed in Subsection \ref{subsection:LIME}, selects a feature, such as a word, from the original input text data and generates many perturbations around that feature by randomly removing or replacing other features (i.e., other words). LIME trains a simpler and explainable model using the perturbed data to generate feature importance scores for each word in the original input text~\cite{ribeiro2016should}. These scores indicate the contribution of each word to the black-box model prediction. LIME identifies and highlights the important words to indicate the impact of the model's prediction, as shown in Figure~\ref{fig:LIME_NLP}.

\textbf{\textit{SHAP:}}
SHAP~\cite{lundberg2017unified} is a widely-used XAI technique in NLP such as text classification, sentiment analysis, topic modeling, named entity recognition, and language generation \cite{ribeiro2016model, jain2019attention}. SHAP computes the feature importance scores by generating a set of perturbations that remove one or more words from the input text data. For each perturbation, SHAP computes the difference between the expected model output when the word is included or not included, which is known as the Shapley value. This approach then computes the importance of each word in the original input text by combining and averaging the Shapley values of all the perturbations. 
Finally, SHAP visualizes the feature importance scores to indicate which words are more useful in the model prediction process.

\textbf{\textit{LRP:}}
To apply LRP to NLP, one must encode preprocessed input text as a sequence of word representations, such as word embeddings, and feed them to a neural network \cite{gholizadeh2021model}. 
The network processes the embeddings using multiple layers and produces the model's prediction. LRP then computes the relevance scores by propagating the model's output back through the network layers. The relevance score for each word is normalized using the sum of all relevance scores and multiplied by the weight of that word in the original input text. This score reflects its contribution to the final prediction.

\textbf{\textit{Integrated Gradients:}}
Integrated Gradients are used in NLP tasks such as text classification, sentiment analysis, and text summarization \cite{pmlr-v70-sundararajan17a}. The Integrated Gradients technique computes the integral of the gradients of the model's prediction with the corresponding input text embeddings along the path from a baseline input to the original input. The baseline input is a neutral or zero-embedding version of the input text that has no relevant features for the prediction task. The difference between the input word embeddings and the baseline embeddings is then multiplied by the integral to find the attribution scores for each word, which indicate the relevance of each word to the model's prediction. Integrated Gradients output a heatmap that highlights the most important words in the original input text based on the attribution scores \cite{shrikumar2017learning}. This provides a clear visualization of the words that were most significant for the model's prediction, allowing users to understand how the model made that particular decision. IG can be used to identify the most important features in a sentence, understand how the model's predictions change when different input features are changed, and improve the transparency and interpretability of NLP models \cite{montavon2017explaining}.

\subsubsection{Prompt-Based Explainability for Transformer Models}
In this subsection, we discuss some of the most common prompt-based explanation techniques, including Chain of Thought (CoT), In-Context Learning (ICL), and interactive prompts.

\textbf{\textit{Chain of Thought:}}
In the context of a large language model (LLM) such as GPT-3\cite{brown2020language}, Chain of Thought prompts refer to the input sequences intended to instruct the model using a series of intermediate reasoning steps to generate a coherent output \cite{wei2022chain, white2023prompt}. This technique helps enhance task performance by providing a clear sequence of reasoning steps, making the model’s thought process more understandable to the audience \cite{jie2024interpretable}. The gradient-based studies explored the impact of change-of-thought prompting on the internal workings of LLMs by investigating the saliency scores of input tokens \cite{wu2023analyzing}. The scores are computed by identifying the input tokens (words or phrases) and inputting them into the model to compute the output. The influence of each token is then calculated through backpropagation, utilizing the gradients of the model. The score reveals the impact of each input token on the model's decision-making process at every intermediate step. By analyzing the step-by-step intermediate reasoning, users can gain a better understanding of how the model arrived at its decision, making it easier to interpret and trust the model's outputs. 

Perturbation-based studies on Chain of Thought explanation through the introduction of errors in few-shot prompts have provided valuable insights into the internal working mechanism behind large language models \cite{madaan2022text, wang2022towards}. Counterfactual prompts have been suggested as a method of altering critical elements of a prompt, such as patterns and text, to assess their impact on the output \cite{madaan2022text}. The study demonstrated that intermediate reasoning steps primarily guide replicating patterns, text, and structures into factual answers. 
Measuring the faithfulness of a CoT, particularly within the context of LLMs, involves assessing the accuracy and consistency with which the explanations and reasoning process align with established facts, logical principles, and the predominant task objectives \cite{lanham2023measuring}. Several key factors are crucial when evaluating CoT faithfulness, including logical consistency, factuality, relevance, completeness, and transparency \cite{lanham2023measuring}. The assessment often requires qualitative evaluations by human judges and quantitative metrics that can be automatically calculated. The development of models to measure faithfulness and the design of evaluation methods remains an area of active research.

\textbf{\textit{Explaining In-Context Learning:}}
In-context learning is a powerful mechanism for adapting the model's internal behavior to the immediate context provided in the input prompt. ICL operates by incorporating examples or instructions directly into the prompt, guiding the model toward generating the desired output for a specific task. This approach enables the model to understand and generate responses that are relevant to the specified task by leveraging the contextual prompts directly from the input. Several studies have focused on the explainability of how in-context learning influences the behavior of large language models, applying various techniques and experimental setups to elucidate this process. A recent study explores a critical aspect of how ICL operates in large language models, focusing on the balance between leveraging semantic priors from pre-training and learning new input-label mappings from examples provided within prompts \cite{wei2023larger}. The study aims to understand whether the LLMs' capability to adapt to new tasks through in-context learning is primarily due to the semantic priors acquired during pre-training or if they can learn new input-label mappings directly from the examples provided in the prompts. The experimental results revealed nuanced capabilities across LLMs of different sizes.

Larger LLMs showed a remarkable capability to override their semantic priors and learn new, contradictory input-label mappings. In contrast, smaller LLMs rely more on their semantic priors and struggle to learn new mappings through the flipped labels. This learning capability demonstrates symbolic reasoning in LLMs that extends beyond semantic priors, showing their ability to adapt to new, context-specific rules in input prompts, even when these rules are completely new or contradict pre-trained knowledge. Another study explores the workings of ICL in large language models by employing contrastive demonstrations and analyzing saliency maps, focusing on sentiment analysis tasks \cite{li2023towards}. In this research, contrastive demonstrations involve manipulating the input data through various approaches, such as flipping labels (from positive to negative or vice versa), perturbing input text (altering the words or structure of the input sentences without changing their overall sentiment), and adding complementary explanations (providing context and reasons along with the input text and flipped labels). Saliency maps are then applied to identify the parts of the input text that are most significant to the model’s decision-making process. This method facilitates visualization of the impact that contrastive demonstrations have on the model’s behavior. The study revealed that the impact of contrastive demonstrations on model behavior varies depending on the size of the model and the nature of the task. This indicates that explaining in-context learning's effects requires a nuanced understanding that considers both the model's architectural complexities and the specific characteristics of the task at hand.

ICL allows large language models to adapt their responses to the examples or instructions provided within the input prompts. Explainability efforts in LLMs aim to reveal how these models interpret and leverage in-context prompts, employing various techniques, such as saliency maps, contrastive demonstrations, and feature attribution, to shed light on LLMs’ decision-making processes. Understanding the workings of ICL in LLMs is crucial for enhancing model transparency, optimizing prompt design, and ensuring the reliability of model outputs across various applications.

\textbf{\textit{Explaining Interactive Prompt:}}
Explaining Interactive Prompt is a technique that focuses on designing and using prompts to interact effectively with large language models~\cite{white2023prompt,slack2023explaining}. This method involves designing prompts that dynamically direct the conversation toward specific topics or solicit explanations. Through the use of strategically designed prompts, users can navigate the conversation with a model to achieve more meaningful and insightful interactions, enhancing the understanding of the model's reasoning and decision-making process. 

Several studies use various approaches to analyze and enhance the effectiveness of explaining interactive prompts. A study called TalkToModel introduced an interactive dialogue system designed to explain machine learning models understandable through natural language conversations or interactive prompts~\cite{slack2023explaining}. It evaluates the system's language understanding capabilities, increasing deeper and more meaningful interactions between users and models through interactive prompts. This approach enhances the interpretability of complex machine learning models' behaviors and the model’s decision-making process. The study called Prompt Pattern Catalog introduced a catalog designed to enhance prompt engineering by systematically organizing and discussing various strategies for constructing prompts \cite{white2023prompt}. This catalog aims to explain the decision-making process of models more clearly. It provides insights and methodologies for eliciting detailed, accurate, and interpretable responses from models, thus improving the understanding of model behavior and decision-making logic.

\subsubsection{Attention Mechanism}

Attention mechanisms in the Transformer architecture enable a model to focus selectively on different parts of the input sequence for each step of the output sequence, mimicking the way humans pay attention to specific parts of a sentence \cite{vaswani2017attention}. Attention weights can be visualized to gain insights into the model's decision-making process, revealing which parts of the input it considers significant. This visualization aids in understanding how the model makes its decisions and explains the significance assigned to various input segments, thereby enhancing the interpretability and transparency of black-box models. AttentionViz \cite{yeh2023attentionviz}, a visualization method for self-attention, highlights query and key embeddings, enabling global pattern analysis across sequences and revealing deeper insights into Transformers' pattern identification and connection beyond previous visualization methods. Another study introduces a method for explaining predictions made by Transformer-based models, particularly those using multi-modal data and co-attention mechanisms \cite{chefer2021generic}. This method provides generic explainability solutions for the three most common components of the Transformer architectures: pure self-attention, a combination of self-attention and co-attention, and encoder-decoder attention.

\subsection{Explainability in Computer Vision}
\begin{figure}[h]
\centering
\includegraphics[width=\linewidth]{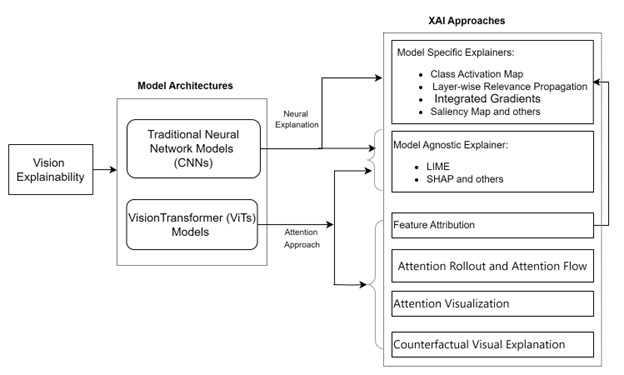}
\caption{Taxonomy of explainability in computer vision. }
\label{fig:vision_taxonomy}
\end{figure}

In computer vision, models can be categorized into Convolutional Neural Network-based models (CNNs) and attention-based models, such as Vision Transformers (ViTs), based on their architecture. Accordingly, various XAI approaches are designed to work with these model architectures. For CNNs, techniques such as saliency maps, LRP, Integrated Gradients, and CAM are the most commonly employed.  
On the other hand, for ViTs, methods like attention visualization, Attention Rollout, Attention Flow, Counterfactual Visual Explanations, and Feature Attribution are extensively used.  Figure~\ref{fig:vision_taxonomy} presents a taxonomy of vision explainability. 

\subsubsection{Explainability of CNNs}
\textbf{\textit{Saliency Maps:}} In computer vision, a saliency map helps identify the most important regions of an image for a deep learning model's prediction. Various methods have been proposed to obtain a saliency map, including deconvolutional networks, backpropagation, and guided backpropagation algorithms \cite{simonyan2013deep, zeiler2014visualizing,springenberg2014striving}.
To generate the saliency map, a pre-processed input image is fed into a pre-trained CNN model to obtain the probability distribution over various classes of the trained model. The output probability gradients are then computed using the backpropagation approach, with higher gradient values indicating the relevance of each pixel to the model's prediction. The saliency map detects the highest gradient value pixels as the most important for the model's prediction, generating a heatmap that highlights these regions in the input image. The resulting saliency heatmap provides important insights into the CNN and aids in interpreting its decision-making processes. The saliency map method removes less relevant regions (pixels), such as the image background, and identifies the most important regions of the input image for the model's decision. However, it is important to note that the saliency map only provides a local explanation by highlighting specific pixels of the input image and does not provide a global explanation.

\begin{figure}[!h]
\centering
\includegraphics[width=\linewidth,scale=0.45]{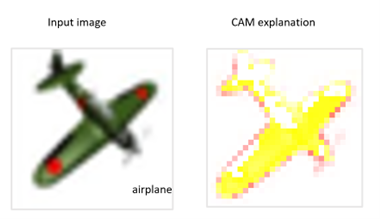}
\caption{Visualization of CAM (input image from CIFAR 10 dataset).}
\label{fig:VisualizationOfCAM}
\end{figure}

\textbf{\textit{Class Activation Maps:}}
CNNs are powerful neural models for image processing tasks, achieving state-of-the-art performance in various applications such as object recognition, segmentation, and detection \cite{krizhevsky2017imagenet, he2016deep}. However, their complex architecture and the high dimensionality of their learned features make it challenging to understand how they make decisions. 
Using CAM for explaining the behavior of CNN models is popular \cite{yang2016wider}. The CNN model is first trained with pre-processed and labeled image data for image classification tasks. A weighted feature map is obtained by multiplying the feature map from the final convolutional layer with channel importance, which highlights the important regions of the input image. The weighted feature map is then passed through a ReLU activation function to keep only positive values. The resulting positively weighted feature map is up-sampled to match the size of the input image. Finally, CAM provides a visual output, as shown in Figure~\ref{fig:VisualizationOfCAM}, by highlighting the most important regions of the original input image using this up-sampled feature map \cite{zhou2016learning,yang2019towards}. CAM does not give specific information on a particular pixel and how that pixel is important to the model's prediction. Nevertheless, this technique can be a valuable tool for interpreting CNNs and enhancing their interpretability. Conveniently, it can additionally be used to improve the robustness of the model by identifying the parts of the images that are irrelevant to the prediction and discarding them \cite{linardatos2020explainable}.

\textbf{\textit{Layer-wise Relevance Propagation:}} 
LRP computes the relevance scores for each pixel in an image by propagating a classification model's output back through the network layers. The relevance score determines the contributions of each pixel to the final model's prediction \cite{bach2015pixel, smilkov2017smoothgrad}.
LRP visualizes the weighted relevance scores as a heat map and highlights pixels in the input image to indicate the most important pixels for the model's prediction. The primary advantage of LRP for image classification tasks is its ability to assign a relevance score for each pixel from the input image, allowing us to understand how much each pixel contributed to the model's output. LRP provides valuable information to interpret the model, offering insights into which pixels are most important for the final prediction. By analyzing the relevance score, LRP helps interpret and understand how the model made the decision and produced the output. Additionally, LRP provides insight into each internal layer's inner workings, offering valuable insights into understanding the internal workings of DNN and improving the model's performance.

\subsubsection{Explainability of Vision Transformers}
Vision Transformers (ViTs) are a category of deep learning models that adapt the Transformer architecture to the domain of image recognition tasks \cite{dosovitskiy2020image}. These models handle images as sequences of patches and treat each patch similarly to how tokens are treated in NLP.

\textbf{\textit{Feature Attribution Techniques:}}
Gradient-based Saliency Maps, Integrated Gradients, CAM, and LRP are among the most common feature attribution techniques employed to explain the decision-making process of complex models such as CNNs and ViTs. Saliency Maps identify the pixels most significant to the model's output \cite{zeiler2014visualizing}. Integrated Gradients trace the contribution of each feature from a baseline to the actual input \cite{pmlr-v70-sundararajan17a}. LRP backpropagates the output decision to assign relevance scores to individual input features \cite{chefer2021transformer}. CAM provides a visual representation of which regions in the input image were most relevant to a particular class by highlighting them \cite{krizhevsky2017imagenet}. Thus, all these approaches provide insights into the model's focus areas for specific features or regions. 

\textbf{\textit{Attention Rollout and Attention Flow Methods:}}
These are Transformer explainability approaches designed for Transformer-based models \cite{abnar2020quantifying}. They are designed to track and understand the flow of information from individual input features through the network's self-attention mechanism. These approaches help interpret the complexity of ViTs by addressing layer-wise information mixing. They provide deeper insights into the decision-making process of the model by offering improved methods for understanding how information from input features is processed and integrated across the network's layers.

\textbf{\textit{Attention Visualization:}}
Transformer Interpretability Beyond Attention Visualization is an advanced Transformer explainability technique that interprets the decision-making process of Transformers, including ViT  \cite{chefer2021transformer}. This method extends beyond traditional attention mechanism visualizations to reveal the nuanced roles of individual neurons, the strategic influence of positional encodings, and the intricate layer-wise interactions within the Transformer. Examining the comprehensive functionalities and contributions of various transformer components aims to provide a more complete explanation of the model's behavior than attention weights alone can provide. This holistic approach enhances our understanding and interpretability of how these models process input features and make decisions.

\textbf{\textit{Counterfactual Visual Explanations:}}
Counterfactual Visual Explanations (CVE) is a vision interpretability technique designed to explain the decision-making processes of ViT models. This method involves changing specific parts of an input image and observing the subsequent changes in the model's output \cite{verma2020counterfactual, guidotti2022counterfactual}. This allows people to identify which image features are most significant in the model's decision-making process. CVE is a practical and insightful approach to understanding how ViT models process and interpret images, providing a tangible means to explore and improve these complex models' interpretability.

\subsubsection{Model-agnostic Explainer for Vision Models}
Model-agnostic approaches, such as LIME and SHAP, have also been adapted to approximate ViTs' behavior with more interpretable models, which is particularly useful for individual predictions. This adaptation process requires considerations such as the computational complexity due to the Transformer structure, which may affect the feature attribution, the high dimensionality of image data, the dependency between input features in image data, which violates the independence assumption in SHAP, and the choice of an appropriate surrogate model that balances fidelity to the original model with interpretability.

\subsection{Explainability in Time Series}
Time series forecasting models are widely used in various domains, such as business, finance, meteorology, and medicine, to predict the future values of a target variable for a given entity at a specific time \cite{shumway2000time,lim2021time}. Time series data refer to an ordered sequence of collected data points accumulated during regularly spaced time intervals. The predictive power of time series forecasting models is rooted in statistical and ML techniques that analyze historical data. However, these models can be complex, unintuitive, and opaque, requiring the use of XAI to ensure fairness and trustworthiness, as well as to improve their accuracy \cite{verma2020time}. The XAI techniques can be used to identify bias in the data, improve the accuracy of predictions, and make the model more interpretable. For instance, XAI can be used to identify whether the model is biased against certain groups of patients, improve the accuracy of the model's predictions by identifying the most important features for its predictions, and provide explanations for the model's predictions.

The use of time series forecasting models is ubiquitous in real-world datasets. Accurate time series forecasting of stock prices, for example, can inform investment decisions and aid in risk management \cite{bao2017deep}. In meteorology and climate science, accurate time series forecasting can aid in weather prediction and disaster risk management \cite{huntingford2019machine}. Devices, such as sensors, accurately record data points with time signatures for use in a combination of statistical and ML techniques to analyze historical data and make forecasts about future values. However, these models face challenges due to outliers, seasonal patterns, temporal dependencies, and missing data, among other factors. Therefore, the quality of the data and model architecture choices play a vital role in the accuracy of time series predictions, necessitating the use of XAI to provide human understanding of the models' performance.

\subsubsection{Saliency Maps}
Generating saliency maps for time series data involves several essential preprocessing steps. Normalization is employed to ensure that all features are on the same scale, facilitating fair comparisons between them. Reshaping transforms the data into a structured representation, where each row corresponds to a time step, and each column represents a feature. Windowing and padding are crucial to capture local relationships within the data \cite{farahat2019convolutional}, enabling the model to discern patterns and dependencies. The resulting saliency map assigns a value to each time step in the input time series, with higher values indicating greater model reliance on that particular time step. This visualization aids in comprehending the interplay among various features and their impact on the final prediction~\cite{huber2021local}.

Saliency maps face challenges in accurately identifying feature importance over time and sensitivity to the underlying model architecture~\cite{ismail2020benchmarking}. Although the concept of saliency maps is model-agnostic, their specific implementation tends to be model-specific, which underscores their role as a valuable output standard for other XAI techniques~\cite{cooper2022believe}. In analyzing time series data, saliency maps offer potential insights into the decision-making process, facilitating the identification of data or model-related issues. However, their dependence on model architecture emphasizes the need to complement their usage with other XAI techniques in comprehensive data analysis.

\subsubsection{CAM}
Although originally designed for image data, CAMs \cite{zhou2016learning} can be applied to time series data by treating each time step as a separate ``channel'' in an image, similar to different color channels, such as red, green, and blue \cite{wang2017time}. The time series can be windowed of fixed size and then stacked as different channels in the image. The size of the windows can be chosen based on the length of the time series and the desired level of detail. For example, if the time series is 1,000 time steps long, you could use windows of size 100 time steps. The windows can be overlapped by a certain amount. 

The last convolutional layer is then used to compute a weighted sum of the feature maps, which is upsampled to the original image size to generate the CAM. This technique utilizes the weights from the global average pooling (GAP) layer to identify the segments of the time series, 
which are then used to produce the saliency mapping of the weight vector to the original time series for visualization and easier explainability. In addition to CAM, there are other approaches for visualizing the intermediate activations of a DNN that utilizes convolutional layers and GAP layers before the final output layer~\cite{springenberg2014striving, springenberg2015better}, and the visualization technique called Deconvolutional Networks 
\cite{zeiler2014visualizing}. Deconvolutional Networks can be repurposed for time series data by treating sequential data points as channels, which enables the visualization and comprehension of hierarchical features learned by convolutional layers in a time series-specific context~\cite{song2020representation}.

\subsubsection{TSViz}
TSViz (Time Series Visualization) is a model-agnostic set of visualization techniques that help users explore and comprehend complex time series data, regardless of the type of model used for analysis \cite{siddiqui2019tsviz}. This XAI technique uses various dimensionality reduction techniques, such as principal component analysis (PCA \cite{labrin2020principal}), t-distributed stochastic neighbor embedding (t-SNE \cite{van2014accelerating}), and uniform manifold approximation and projection (UMAP \cite{mcinnes2018umap}), to simplify high-dimensional time series data and generate visualizations that reveal trends, patterns, and correlations \cite{agrawal2021time, roy2020umap}. These techniques are especially useful for identifying complex relationships that might be difficult to detect with traditional visualization methods \cite{munir2021thesis}.

TSViz is a post-hoc, human-in-the-loop XAI tool that supports analysis and decision-making by human experts~\cite{siddiqui2019tsviz}. Human interpretation and input are essential for comprehending the data and visualizations generated, and users are responsible for determining metrics and making necessary adjustments to the model \cite{mosqueira2022human}. TSViz enhances users' ability to analyze and model time-series data through various visualizations, including line plots, heatmaps, seasonal plots, autocorrelation plots, and spectral density plots. However, it does not replace the importance of human expertise and judgement required for effective decision-making and insight drawing \cite{schlegel2021time}. Overall, TSViz is an essential tool that empowers users to make data-driven decisions and predictions, gain a deeper understanding of underlying systems and phenomena, and identify potential sources of bias in their models.

\subsubsection{LIME}
Applying LIME to time series data introduces a unique set of challenges stemming from the inherent temporal dependencies within such data. In contrast to static tabular data, time series data necessitates an understanding of how events evolve over time. 
By crafting local, interpretable models that approximate black-box model predictions within specific time segments, LIME equips analysts with a powerful tool for deciphering the temporal nuances that influence outcomes \cite{plumb2018interpretable}. One way to do this is to use a sliding window of fixed length that moves along the time series data, or to use an attention mechanism that identifies which parts of the time series are most relevant to the prediction at each time point \cite{lipton2016modeling}.

Let's consider a specific example: a prediction model for the stock price on a particular day is explained using LIME~\cite{lakkaraju2016interpretable}. The most important time steps for the prediction is identified by obtaining feature importance scores or time series cross-validation. For example, we may find that the previous day's stock price, trading volume, and news articles are the most important features. 
Next, the time series is perturbed by randomly changing the values of the most important time steps. The amount of change is controlled by a hyperparameter 
to reduce noise in the local model~\cite{ribeiro2016model}. The perturbed time series is fed to the black-box model and records the corresponding predictions. This process is repeated for multiple perturbed instances. Then, we train a local, interpretable model using the perturbed time series and their associated predictions. The coefficients of the model can be examined to identify the most important features. For instance, the previous day's stock price has the highest coefficient, indicating that it is the most important feature for the prediction. Finally, the coefficients of the local model are used to explain the model's prediction. If the coefficients show that the previous day's stock price was the most important feature, we can say that the black-box model predicted the stock price based on the previous day's price.

\subsubsection{SHAP}
SHAP can be seamlessly applied to time series data, where it offers valuable global insights into a model's decision-making process over time. This capability aids in comprehending the evolving behavior of time series models and the identification of trends or patterns influencing predictions. 

First, we train a time series model on a dataset of historical time series data. Once the model is trained, we can use SHAP to explain the output of the model for a new time series. To do this, we need to calculate the SHAP values for each feature in the time series. 
For time series data, each feature corresponds to a specific time step in the time series.
To calculate the Shapley values for a time series, we create a dataset that contains perturbations of the original time series. Each perturbation consists of a modified version of the original time series where the value of one time step is changed while keeping the values of all other time steps fixed. 
These differences are used to calculate the Shapley values for each time step in the time series. The Shapley values can be used to understand how the model works and to identify the most important features for the prediction. For example, we can plot the Shapley values for each time step in the time series to visualize the contribution of each time step to the model's output.

\subsection{Explainability in Healthcare}
Healthcare is a critical domain due to the high risks and complexities of the medical decision-making process. For example, a study has found that human surgeon usually explains the details of the surgery beforehand and has a 15\% chance of causing death during the surgery. A robot surgeon has only a 2\% chance of death. A 2\% risk of death is better than a 15\% risk, but people prefer the human surgeon \cite{rudin2019we}. The challenge is how to make the physicians and patients trust the deployed AI model within the robot. How much communication and explanation between a robot, patient, or physician is sufficient? The power of XAI comes here to provide transparent, understandable, and interpretable explanations for the AI model decisions to build trust between them. Hence, XAI is necessary in healthcare due to its role in enhancing trust and confidence, ensuring faithfulness to ethical considerations, managing regulatory compliance, improving clinical decision-making, facilitating ongoing learning and model fine-tuning, improving risk management and analysis, and enhancing collaborative communications between clinicians, patients, and AI models \cite{hamamoto2021application}. 
XAI has several crucial applications in various aspects of healthcare, such as medical diagnosis, patient treatment, drug discovery and development, clinical decision support, and risk assessment.

\textbf{ \textit{Medical diagnosis:}} Medical data in healthcare are diverse and sophisticated, incorporating various types and sources, including imaging (like MRIs, CT scans, and X-rays), electronic health records, genomic data, laboratory test results, patient-reported data, wearable device data, social and behavioral data, and pharmacological data. XAI improves disease diagnostics by providing transparent interpretations of AI model decisions, ensuring accurate disease identification, and predicting patient outcomes through comprehensible explanations of the model outputs for these complex medical data \cite{albahri2023systematic, bharati2023review}. XAI also helps identify influential features from the complex medical data affecting the model decision-making process \cite{lotsch2021explainable}.

\begin{table*}[h]
  \begin{center}
    \caption{Summarize XAI techniques in Healthcare.}        \label{HealthcareSumXAItechniques}
    \begin{tabular}{m{10pt}|m{40pt}|m{70pt}|m{90pt}|m{105pt}|m{50pt}|m{55pt}} \hline
        \textbf{No.} & \textbf{Data Type}&\textbf{Medical data}&\textbf{XAI Techniques}& \textbf{Application Areas} & \textbf{Benefits} & \textbf{Papers} \\ \hline  
        1 & Image &X-rays, ultrasound, MRI and CT scans, etc&LRP, LIME, CAM, Saliency Maps, and Integrated Gradients &Radiology, Pathology, Dermatology, Ophthalmology, and Cardiology & Interpretable image analysis &\cite{li2019large, bian2019weakly, rajaraman2019visualizing, yang2015manifold} \\ \hline
        
        2& Text  & Clinical text, Electronic Health Records (EHRs), and case studies  & LIME, SHAP, Attention Mechanism, and Counterfactual Explanations &  Drug Safety, and Medical Research & Interpretable text analysis & \cite{ahmed2022eandc, ming2018rulematrix,  rane2023explainable}  \\ \hline
        
        3& Structured (Numeric)  & Patient demographics, laboratory test results, pharmacy records, billing and claims  & LIME, SHAP, Decision Trees, Rule-based Systems, Counterfactual Explanations, Integrated Gradients, and BERT Explanations & Patient Health Monitoring and Management, Epidemiology, and Clinical Trials and Research & Interpretable structured data analysis & \cite{guidotti2018survey, rane2023explainable,magunia2021machine} \\ \hline 
        
        4& Time series  & ECGs, EEGSs, monitoring and wearable device data & TSViz, LIME, CAM, SHAP, Feature Importance, and Temporal Output Explanation  & Neurology and EEG Monitoring, Patient Monitoring in Critical Care, and Cardiology and Heart Health Monitoring & Interpretable time series analysis & \cite{raza2022designing, morabito2023explainable} \\  \hline
        
        5& Multi-modal & Telemedicine interactions (text, audio, video) & LIME, SHAP, Attention Mechanisms, Multi-modal Fusion and Cross-modal Explanations & Cancer Diagnosis and Treatment, Neurology and Brain Research, and Mental Health and Psychiatry & Interpretable multi-modal analysis & \cite{el2021multilayer, yang2022unbox} \\ \hline 
        
        6& Genetic & Genetic makeup & Sensitivity Analysis, LIME, SHAP, and Gene Expression Network Visualization & Genomic Medicine, Oncology, and Prenatal and Newborn Screening & Interpretable genetic analysis & \cite{awotunde2022explainable, anguita2020explainable} \\  \hline 
        
        7&  Audio & Heart and lung sounds & Saliency Maps, LRP, SHAP, LIME, and Temporal Output Explanation & Cardiology, Pulmonology, Mental Health, and Sleep Medicine & Interpretable audio analysis & \cite{troncoso2022explainable, tjoa2020survey} \\  \hline			
\end{tabular}
\end{center}  
\end{table*}

\textbf{ \textit{Patient treatment:}} 
By examining the explanations of unique health data, XAI helps design treatment plans for individual patients. It provides good insights into why a specific medicine is suggested based on a patient's medical data and diagnosis~\cite{liao2023artificial}.

\textbf{ \textit{Drug discovery and development:}} XAI plays a vital role in drug discovery and development in the pharmaceutical industry \cite{askr2023deep}. It explains and provides insights into complex relationships among the molecular structures of drugs and their biological effects \cite{kha2023development}.
    
\textbf{ \textit{Clinical decision support:}} XAI systems assist healthcare professionals by providing transparent and interpretable explanations of the decision-making processes of models handling complex data \cite{panigutti2023co}. Clinicians can more easily understand complex cases by considering the influential features highlighted in the explanations provided by XAI systems~\cite{loh2022application}.

\textbf{ \textit{Legal considerations:}} The use of XAI in healthcare raises several legal considerations. The model decision-making process and how and why the model made that decision should be transparent and understandable for healthcare professionals and patients \cite{albahri2023systematic}. XAI systems should be ensured and safeguarded for the privacy and security of medical and patient data \cite{saraswat2022explainable}. XAI systems mitigate the biases and ensure that AI-made decisions are fair and reasonable across diverse patients \cite{albahri2023systematic}. Regulatory transparency and audits \cite{ward2020machine}, medical device regulations \cite{ma2021understanding}, informed consent \cite{sharma2022artificial}, liability and malpractice concerns \cite{maliha2021artificial}, and intellectual property rights \cite{alam2023explainable} are among the most critical legal considerations in the application of XAI in healthcare.

\textbf{ \textit{Ethical considerations:}} The ethical considerations of XAI in healthcare are complicated and significant \cite{amann2020explainability, chaddad2023survey}. It focuses on the sensitivity and importance of medical information, the decision of the AI model, and the explanations of the deployed XAI system. Transparency and accountability \cite{albahri2023systematic}, fairness and bias mitigation \cite{amann2020explainability}, ethical frameworks and guidelines \cite{kerasidou2021ethics}, privacy and confidentiality \cite{aranovich2023ensuring} are some of the key ethical aspects of XAI in healthcare.
Medical data are complex and diverse, requiring the use of a variety of XAI techniques to interpret it effectively. 
Table~\ref{HealthcareSumXAItechniques} summarizes various XAI techniques and their application areas in healthcare. XAI faces several challenges in healthcare, such as the complexity and diversity of medical data, the complexity of AI models, updating XAI explanations in line with the dynamic nature of healthcare, the need for domain-specific knowledge, balancing accuracy and explainability, and adhering to ethical and legal implications \cite{anton2022comprehensive, alam2023explainable}.

\subsection{Explainability in Autonomous Vehicles}
Autonomous vehicles use complex and advanced AI systems by integrating several deep-learning models that can handle various data types, such as images, videos, audio, and information from LIDAR and radar \cite{al2022generating}. These models utilize inputs from diverse sources, including cameras, sensors, and GPS, to deliver safe and accurate navigation. A crucial consideration is determining which data are most critical. What information takes precedence, and why? Understanding the importance of different data types is key to enhancing our models and learning effectively from the gathered information \cite{alshami2023pose2trajectory}. To address these questions and better decision-making processes of black-box AI models, developers use the XAI approach to evaluate the AI systems. Implementing XAI in autonomous vehicles significantly contributes to human-centered design by promoting trustworthiness, transparency, and accountability. This approach considers various perspectives, including psychological, sociotechnical, and philosophical dimensions, as highlighted in Shahin et al.~\cite{atakishiyev2021explainable}. Table \ref{AutoCarSumXAItechniques} shows a summary of various XAI techniques in autonomous vehicles, including visual, spatial, temporal, audio, environmental, communication, genetic, and textual. The advancements significantly enhance the AI-driven autonomous vehicle system, resulting in a multitude of comprehensive, sustainable benefits for all stakeholders involved, as follows.

\textbf{ \textit{Trust:}} User trust is pivotal in the context of autonomous vehicles. Ribeiro et al.~\cite{ribeiro2016should} emphasized this by stating: ``If users do not trust a model or its predictions, they will not use it''. This underscores the essential need to establish trust in the models we use. 
XAI can significantly boost user trust by providing clear and comprehensible explanations of system processes \cite{holliday2016user}. Israelson et al.~\cite{israelsen2019dave} highlighted the critical need for algorithmic assurance to foster trust in human-autonomous system relationships, as evidenced in their thorough analysis. The importance of transparency in critical driving decisions, noting that such clarity is crucial for establishing trust in the autonomous capabilities of self-driving vehicles~\cite{atakishiyev2021towards}.

\textbf{ \textit{Safety and reliability:}} These are critical components and challenges in developing autonomous driving technology~\cite{corso2020interpretable}. Under the US Department of Transportation, the American National Highway Traffic Safety Administration (NHTSA) has established specific federal guidelines for automated vehicle policy to enhance traffic safety, as outlined in their 2016 policy document 
~\cite{mcgehee2016review}. In a significant development in March 2022, the NHTSA announced a policy shift allowing automobile manufacturers to produce fully autonomous vehicles without traditional manual controls, such as steering wheels and brake pedals, not only in the USA but also in Canada, Germany, the UK, Australia, and Japan \cite{atakishiyev2021explainable}. Following this, The International Organization for Standardization (ISO) responded by adopting a series of standards that address the key aspects of automated driving. These standards are designed to ensure high levels of safety, quality assurance, efficiency, and the promotion of an environmentally friendly transport system \cite{rahman2022transportation}. Besides, Kim et al.~ \cite{kim2020advisable,kim2021toward} described the system's capability to perceive and react to its environment: The system can interpret its operational surroundings and explain its actions, such as ``stopping because the red signal is on''.

\textbf{ \textit{Regulatory compliance and accountability:}} 
Public institutions at both national and international levels have responded by developing regulatory frameworks aimed at overseeing these data-driven systems \cite{atakishiyev2021explainable}. The foremost goal of these regulations is to protect stakeholders' rights and ensure their authority over personal data. The European Union's General Data Protection Regulation (GDPR) \cite{regulation2016regulation} exemplifies this, establishing the ``right to an explanation'' for users. This principle underscores the importance of accountability, which merges social expectations with legislative requirements in the autonomous driving domain. XAI plays a pivotal role by offering transparent and interpretable insights into the AI decision-making process, ensuring compliance with legal and ethical standards. Additionally, achieving accountability is vital for addressing potential liability and responsibility issues, particularly in post-accident investigations involving autonomous vehicles, as highlighted by Burton et al. \cite{burton2020mind}. Clear accountability is essential to effectively manage the complexities encountered in these situations.

\textbf{ \textit{Human-AI decision-making (collaboration):}} In recent autonomous vehicle systems, machine learning models are utilized to assist users in making final judgments or decisions, representing a form of collaboration between humans and AI systems \cite{chen2023understanding}. With XAI, these systems can foster appropriate reliance, as decision-makers may be less inclined to follow an AI prediction if an explanation reveals flawed model reasoning \cite{bussone2015role}. From the users' perspective, XAI helps to build trust and confidence through this collaboration. In contrast, in terms of developers and engineers, XAI helps to debug the model, identify the potential risks, and enhance the models and the vehicle technology~\cite{dong2023did}.

\begin{table*}[!h]
  \begin{center}
    \caption{Summarize XAI techniques in Autonomous Vehicles.} \label{AutoCarSumXAItechniques} 
    
    \begin{tabular}{m{10pt}|m{55pt}|m{40pt}|m{40pt}|m{65pt}|m{170pt}|m{25pt}} \hline
        \textbf{No.} & \textbf{Input Types}&\textbf{Sources}&\textbf{AI \newline models}& \textbf{XAI \newline Techniques} & \textbf{Key Benefits} &\textbf{Papers} \\ \hline 
        1 & Visual Data  & Camera Images and Video Streams & CNN, ViTs, RNN, or LSTM & LRP, CAM, Saliency Maps, Integrated Gradients, Counterfactual Explanations & Enhancing visual environmental interaction, understanding dynamic driving scenarios, allowing correct object detection, interpreting real-time decision-making, and adaptive learning process by providing insights into AI’s model prediction. & \cite{mankodiya2022od, karim2022toward} \\ \hline 
        
        2 & Spatial Data  & LIDAR and Radar & CNN, DNN & Feature Importance, SHAP, CAM, LRP, LIME & Enhancing 3D space and object interactions, improving safety, security, reliability, design, development, and troubleshooting of the car by providing insights into the AI's spatial data processing and model decision-making & \cite{madhav2022explainable, onyekpe2022explainable} \\ \hline 
        
        3 & Temporal Data (Time-Series) & Sensor & RNN, LSTM, GRU & TSViz, LIME, CAM, SHAP, LRP Feature Importance, Temporal Output Explanation & provides insights into time-series data for reliable decision-making, identifying potential safety issues, enhancing overall vehicle safety, and a deeper interpretation of the vehicle's actions over time for post-incident analysis & \cite{cheng2021method, rojat2021explainable}\\ \hline 
        
        4 & Auditory Data  & Micro-\newline phone  & CNN, RNN & LRP, CAM, Saliency Maps, Attention Visualization & enhancing the vehicle's ability to understand and react to auditory signals, improving safety and security by providing insights into AI's audio data decision-making process & \cite{atakishiyev2021explainable} \\ \hline 
        
        5 & Environmental Data (Weather \& Geolocation) & GPS and Cloud-based services & GNN, Random Forest, Gradient Boosting & Rule-based Explanations, Decision Trees, SHAP, LIME, Counterfactual Explanations & Enhanced decision-making, improving safety and efficiency, increasing safety, security, and reliability by providing insights into environmental factors and diverse conditions in the AI model’s decision-making process & \cite{atakishiyev2021explainable,nwakanma2023explainable} \\ \hline 
        
        6 & Vehicle Telematics (engine \& internal status) & Engine Control Unit, On-Board Diagnostics, Sensors & DNN, SVM & LRP, LIME, SHAP, Decision Trees, Rule-based Explanations, Counterfactual Explanations & Helping to interpret engine data and vehicle status information, predicting maintenance and potential issues, improving safety, reducing risk, and providing clear vehicle health status through insights gained from the AI model decision-making process  & \cite{li2023intelligent} \\ \hline 
        
        7 & Communica-\newline tion Data (Vehicle-to-Everything or V2X) & Personal devices, Cloud, Vehicles, Cellular networks & Reinfor-\newline cement Learning & LRP, Saliency Maps, Counterfactual Explanations & Provide insights into model decision-making that enhances trust and safety, interactions with external factors, and improving decision-making by interpreting complex V2X communications  & \cite{bendiab2023autonomous} \\ \hline 
        8 & All & All & All & Generative Language Model Explanation & Provides textual explanations to model users that enable them to interpret  diversified datasets and complex AI model decision-making process & \cite{dong2023did} \\ \hline

\end{tabular}
\end{center}  
\end{table*}

\subsection{Explainability in AI for Chemistry and Material Science}
In chemistry and material science, AI models are becoming increasingly sophisticated, enhancing their capability to predict molecular structures, chemical reactions, and material behaviors, as well as discover new materials \cite{maqsood2024future}. Explainability in chemistry and material science increases beyond simply understanding and analyzing model outputs; it encompasses understanding the rationale behind the model predictions \cite{oviedo2022interpretable}. XAI techniques play a crucial role in obtaining meaningful insights and causal relationships, interpreting complex molecular behaviors, optimizing material properties, and designing innovative materials through the application of AI models \cite{pilania2021machine}. By explaining how and why machine learning models make predictions or decisions, researchers and practitioners in the field can more confidently trust machine learning models for analytical investigations and innovations.  This understanding is important to increasing trust, facilitating insights and discoveries, enabling validation and error analysis, and dealing with regulatory and ethical considerations in AI models \cite{choudhary2022recent}.  The study, ``CrabNet for Explainable Deep Learning in Materials Science''~\cite{wang2022crabnet}, focuses on improving the compositionally restricted attention-based network to produce meaningful material property-specific element representations. These representations facilitate the exploration of elements' identities, similarities, interactions, and behaviors within diverse chemical environments \cite{wang2022crabnet}. Various model-agnostic and model-specific interpretability methods are employed in chemistry and material science to explain the prediction of black-box models' molecular structure, chemical reactions, and the relationship between chemical composition \cite{lee2022comparison, feng2020explainable, harren2022interpretation} and design of new materials \cite{pilania2021machine}. 

\subsection{Explainability in Physics-Aware AI}
Physics-aware artificial intelligence focuses on integrating physical laws and principles into machine learning models to enhance the predictability and robustness of AI models \cite{willard2020integrating}. Explainability in physics-aware AI is crucial for understanding and interpreting these models. It also bridges the gap between the black-box nature of AI models and physical understanding, making them more transparent and trustworthy \cite{datcu2023explainable}. Several approaches exist to explain physics-aware AI models \cite{willard2022integrating}. Domain-specific explanation methods are designed for specific domains, such as fluid dynamics, quantum mechanics, or material science \cite{huang2022physically, crocker2023using}. Model-agnostic explanations are also used to explain the general behavior of physics-aware AI models to ensure their decisions are understandable in various scenarios regardless of the specific model architecture \cite{sadeghi2023explainable, roscher2020explainable}. In the context of physics-aware AI, explainability offers several key advantages, such as enhancing trust and interpretability, ensuring physically plausible predictions, improving model architecture and debugging, providing domain-specific insights, bridging knowledge gaps, ensuring regulatory compliance, and facilitating human-AI collaboration \cite{datcu2023explainable, tuia2023artificial}.

\section{XAI Evaluation Methods}
\label{Evaluation}
\begin{figure}[h]
\centering
\includegraphics[width=\linewidth]{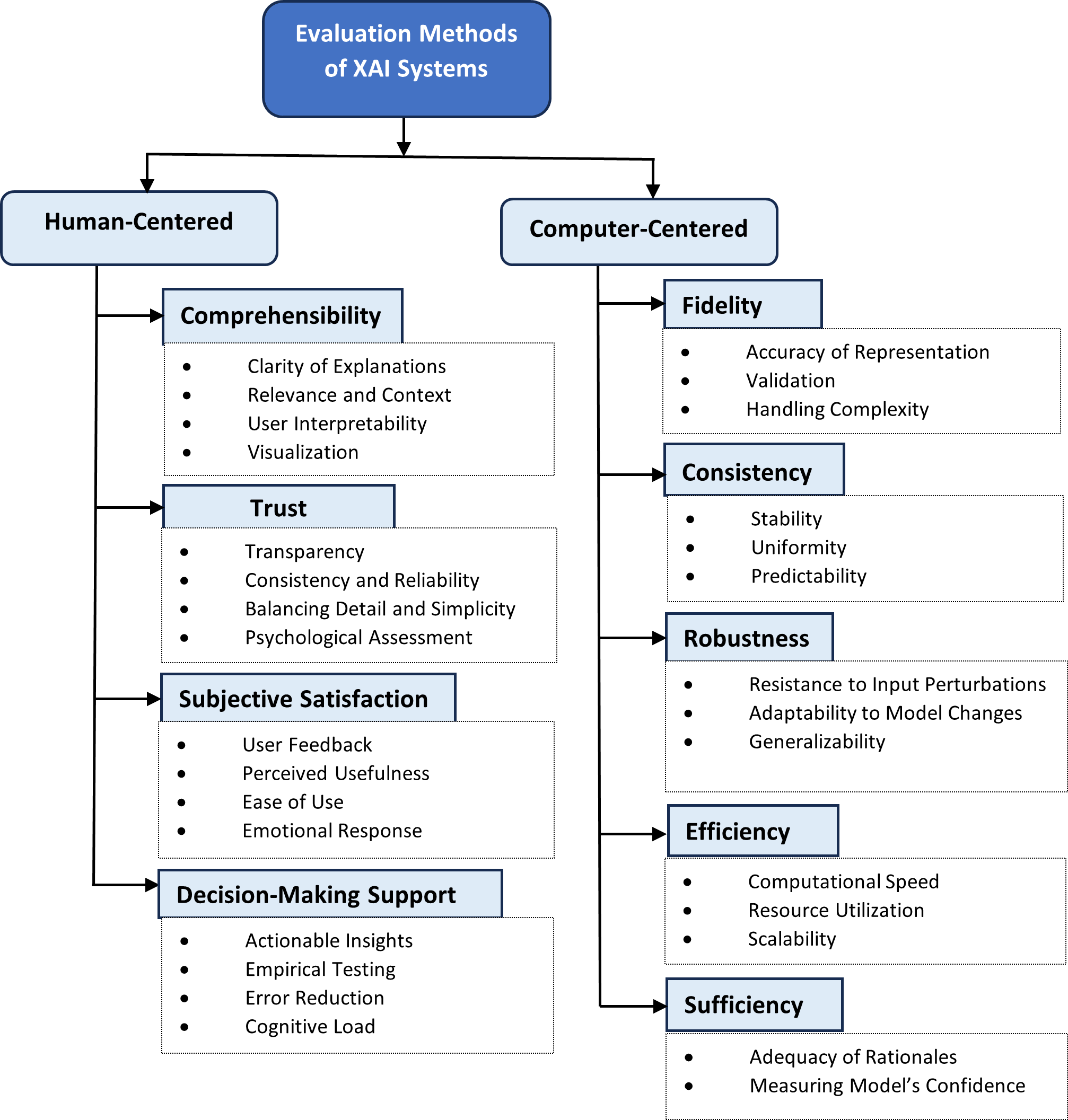}
\caption{A suggested classification framework for assessing the efficacy of XAI systems (adapted from \cite{lopes2022xai}).}
\label{fig:EvaluationMethod}
\end{figure}

XAIs are essential in today’s advancing AI world to ensure trust, transparency, and understanding of AI ethical decision-making, particularly in sensitive domains like healthcare, finance, military operation, autonomous systems, and legal issues. However, we need evaluation mechanisms to measure the generated explanations to ensure their quality, usefulness, and trustworthiness. XAI system evaluation methods are classified into human-centered and computer-centered categories based on their applications and methodologies to judge the effectiveness of XAI techniques \cite{lopes2022xai, hassija2024interpreting}. Figure~\ref{fig:EvaluationMethod} shows a simple taxonomy of XAI evaluation methods.

\subsection{Human-Centered Approach}
The human-centered approach evaluates how the provided XAI explanations meet model users' needs, understanding levels, and objectives from the human perspective \cite{mohseni2021multidisciplinary}. This approach is concerned with comprehensibility, trust, user satisfaction, and decision-making support of the provided XAI explanations \cite{mohseni2018human}. The generated explanations should be clear, brief, and easily understandable by the end-users without technical knowledge.  The explanation should also be relevant and represented by a clear and interpretable visualization mechanism. The explanations should build the end-users' trust by providing a transparent, consistent, and reliable decision-making process for the model \cite{gunning2019darpa}. Hence, the designed evaluation methods ensure the users' trust and satisfaction level through surveys, interviews, questionnaires, behavioral analysis, and other handy tools. The users’ satisfaction is the most essential aspect of evaluation methods. The XAI system can be evaluated by collecting the users' feedback and assessing their emotional responses~\cite{nourani2019effects}. Ease of using the XAI system and the usefulness of the generated explanations to end users also provide insights into the values of the explanation and XAI system. XAI systems can be assessed by evaluating how effectively the generated explanations support users' decision-making. How much do the XAI explanations apply to their decision-making process, reduce errors, and enhance productivity? The human-centered evaluation method also assesses the cognitive load that ensures the provided XAI explanations do not affect the user's cognitive processing capacity \cite{gunning2019darpa}.

\subsection{Computer-Centered Approach}
The computer-centered method aims to the evaluation of XAI techniques based on technical and objective standards without the interactions of human interpretation \cite{hedstrom2023quantus}.
This method involves important XAI technical, objective, and quantifiable metrics such as fidelity, consistency, robustness, efficiency, and other dimensions \cite{zhou2021evaluating}. 

\subsubsection{Fidelity}
Fidelity refers to how the provided XAI explanations are close to the actual decision made by a model focusing on the accuracy of representation, quantitative measurement, and complex model handling \cite{markus2021role}. Does the explanation reflect the accurate reasoning decision process of a model? Does the explanation contain essential information about complex models, such as deep learning models? Hence, high fidelity reflects that the explanation is an accurate interpretation. Fidelity is computed at the instance level using the following formula \cite{velmurugan2021developing}:

\begin{equation}
S = 1 - \sum_{i=1}^{n} \frac{\left| Y(x_i) - Y(x'_i) \right|}{\left| Y(x_i) \right|},
\end{equation}
where $n$ is total number of inputs, \( x \) is the original input for the process instance, \( X' \) is the set of perturbations for \( x \) and \( x' \in X' \), \( Y(x) \) is the model output given input \( x \), and \( Y(x') \) is the model output given input \( x' \).

\subsubsection{Consistency}
Consistency focuses on the stability and coherence of the explanations provided by the XAI system with the same input and different model runs. Stability, uniformity, and predictability are the most important aspects of consistency. The consistency of XAI systems can be determined in different ways \cite{sun2015stability}. The following equations compute stability and uniformity.

Stability can be represented by the variance in explanations over multiple runs with the same input.
\begin{equation}
\sigma^2_{\text{exp}} = \frac{1}{N}\sum_{i=1}^{N} (e_i - \bar{e})^2,
\end{equation}
where \( \sigma^2_{\text{exp}} \) is variance of explanations, \( e_i \) is the explanation for the $i^{th}$ run, \( \bar{e} \) is the average of all explanations, and \( N \) is the total number of runs.

Uniformity ($U$) quantifies how uniformly distributed the explanation features are. The higher value of $U$ implies that the distribution of features in the explanation is closer to a uniform distribution.
\begin{equation}
\text{U} = 1 - \sqrt{\frac{1}{N}\sum_{N=1}^{N} \left(r_n - \frac{1}{N}\right)^2},
\end{equation}
where $r_n$ is the relevance of $n^{th}$ feature.

\subsubsection{Robustness}
The robustness of the XAI system assesses the resilience of explanations in the behavior of model change in various aspects such as input change and adversarial attack. The robustness also focuses on model update adaptability and generalizability. Is the evaluation method applicable to various XAI systems on various platforms?  Does the evaluation method continue its function if the XAI system is updated? The robustness of XAI systems can be computed using the perturbation approach using the following formulas \cite{drenkow2021systematic}. 

Resistance to Input Perturbations ($R$), obtained using the following equation: the change in explanations with slightly perturbed inputs. 
\begin{equation}
 R = 1 - \frac{1}{N}\sum_{i=1}^{N} \| exp(x_i) - exp(x'_i) \|,
 \end{equation}
where $exp(x)$ is the explanation for input \( x \), \( x'_i \) is the perturbation of \( x_i \), and \( N \) is the total number of perturbations.

Adaptability to Model Changes ($A$): the change in explanations after the model is updated. 
\begin{equation}
 A = \frac{1}{N}\sum_{i=1}^{N} \| exp_m(x_i) - exp_{m'}(x_i) \|,
 \end{equation}
where $exp_m(x)$ is the explanation from model $m$, $exp_{m'}(x)$ is from the updated model $m'$, and $N$ is the total number of samples.

\subsubsection{Efficiency}
The efficiency of the evaluation method involves computational capacity and resources, such as resource utilization and time, to generate explanations and scalability that the evaluation method handles large-scale explanations. The following simple computing formulas can represent computational speed and scalability, respectively \cite{schryen2022speedup}.

Computational speed ($C_{s}$) is the rate at which an XAI system can generate explanations. 
\begin{equation}
C_s = \frac{1}{T \times R},
\end{equation}
where $T$ is the time taken to generate an explanation, and 
$R$ is the computational resources used, such as memory or CPU cycles. The lower $C_{s}$ indicates the higher efficiency.

Scalability ($S$) is the ability of an XAI system to handle increasing volumes of input data and explanations.
\begin{equation}
S = \lim_{n \to \infty} \frac{P(n)}{n},
\end{equation}
where $P$ is the system's performance measure as the input $n$ size grows. The XAI system is scalable and efficient if $S$ is bounded as $n$ increases.

\subsubsection{Sufficiency}
The sufficiency metric assesses the adequacy of rationales in supporting the model's decision-making process \cite{deyoung2019eraser}. A rationale is a subset of input data that a model identifies as critical for its decision-making process. This metric evaluates whether the rationales alone are sufficient for the model to maintain its prediction confidence. It measures the difference in the model’s confidence between using the entire input and using just the rationale, and it is mathematically represented as:
\begin{equation}
\text{sufficiency} = m(x_i)_j - m(r_i)_j
\end{equation}
where $m(x_i)_j$ is the model's confidence score for the full input $x_i$ and  $m(r_i)_j$ is the model's confidence score when only the rationale $r_i$ is provided.\\
Measuring the difference in prediction confidence between the full input and the rationale helps determine whether the rationale alone can sustain the model’s decision-making process. A small difference indicates high sufficiency, meaning the rationale captures most of the essential information. In contrast, a large difference suggests low sufficiency, indicating that the rationale may be missing important information or that the model relies on other parts of the input.

We can use alternative formulas to compute sufficiency.\\
\textbf{ \textit {Confidence Ratio (CR):}} It calculates the ratio of the model's confidence with the rationale to the confidence with the full input.

\begin{equation}
\text{CR} = \frac{m(r_i)_j}{m(x_i)_j}
\end{equation}
A CR close to 1 indicates the rationale is highly sufficient, a significantly lower CR (close to 0) suggests it is insufficient, and a CR greater than 1 might indicate potential overfitting or anomalies in the model's reliance on the rationale.\\
\textbf{ \textit {Percentage Drop in Confidence (PDC):}}  It
 measures the percentage decrease in confidence when using the rationale compared to the full input.

\begin{equation}
\text{PDC} = \left(1 - \frac{m(r_i)_j}{m(x_i)_j}\right) \times 100\%
\end{equation}
A lower PDC indicates that the rationale is highly sufficient with minimal loss of confidence, whereas a higher PDC suggests that the rationale is insufficient.

\section{Future Research Direction}
\label{FeatureReserechDirections}
 The existing XAI systems have faced several challenges in various aspects, including design objectives, applications, standardization, model complexity, security, and evaluation metrics. Further research is required to overcome these challenges and enhance the state-of-the-art. In this section, we discuss some of the most common XAI challenges and future research directions.

 \subsection{Model Complexity}
 XAI techniques are often less effective with highly complex models \cite{das2020opportunities}. Developing AI models by reducing complexity without compromising much on accuracy is a challenging task. Model simplification, building hybrid models, and interactive explanations may be possible approaches to overcoming the existing challenges.

\subsection{Building ML Models with Explanation }
Building AI models with explanations is crucial for safety-critical application areas \cite{thampi2022interpretable}. However, it is not only a technical challenge but also involves practical considerations such as ethical and legal issues. Building accurate AI models with explanations is complex and requires further research. Applying XAI in the training stage, using data-driven insights, understanding the model's predictions, and continuous interpretations may be the possible approaches to building AI models with explanations \cite{dwivedi2023explainable}. 

\subsection{Performance vs. Interpretability}
The trade-off between Performance and Interpretability is one of the biggest challenges. Simplifying a model for better interpretability may lead to reduced accuracy of the model. Performance is critical for time-sensitive applications and complex models, whereas interpretability and explainability are essential for safety-critical applications to trust a model. Hence, the recommended solution is finding a balance between performance and interpretability.

\subsection{Standardization and Evaluation Methods}
The right evaluation metrics are essential to measure the performance of XAI systems. The AI models are designed to solve various problems with various design objectives \cite{lopes2022xai}, and these diversified AI systems require different types of XAI systems. Hence, applying the same evaluation metrics to different XAI systems can be challenging because the design objectives of XAI systems are different. For example, the design objectives of interpretability, accuracy, fairness, robustness, and transparency are different. Each of these XAI design objectives may require different evaluation metrics, and it is essential to select the right evaluation metrics that align with the design objectives of the XAI systems to address this challenge. Moreover, applying the combinations of different evaluation metrics to measure the performance of XAI systems may be helpful.

\subsection{Security and Privacy}
The applications of complex AI systems have exhibited several challenges toward an ethical code, such as security, privacy, fairness, bias, accountability, and transparency. For example, the diversity of ethical issues is one of the main challenges, as current ethical studies have shown. XAI systems help investigate and explain how and why the model made such an ethical decision. However, XAI systems themselves result in privacy, security, and other related challenges and require another special consideration. XAI explanations may be causes of information leakage, model inversion attacks, adversarial attacks, and explanation integrity. There is a trade-off between explainability with security and privacy. Balancing this trade-off using some strategies, including privacy-preserving, selective explanation, anonymization, secure communication, and auditing techniques, is crucial. Integrating differential privacy techniques, generating explanations that only highlight broad features, producing a generalized explanation, and implementing access control mechanisms may also be possible to overcome the trade-off between performance and interpretability.

\subsection{Explainability of Multi-modal Models}
Multimodal AI models are designed to process multiple data modalities, such as texts, images, audio, videos, and other modalities. Explaining the fusion of modalities, intermodal relationships, scalability, data heterogeneity, and high dimensionality are some of the most complex challenges in the current state-of-the-art \cite{chefer2021generic}. Therefore, designing combined XAI techniques may be helpful to address these challenges. The design process requires multi-disciplinary efforts from machine learning, computer vision, natural language processing, and human-computer interaction expertise. Specifically, large language models, GPT, are on the way to becoming Any-to-Any Multimodal models \cite{wu2023next}. Hence, the Any-to-Any Multimodal explanation technique is required, but it is complex and challenging.

\subsection{Real-time Explanation}
Complex AI models have recently become more popular and deployed in real-time situations in various non-safety and safety-critical application areas. Safety-critical application areas such as healthcare monitoring, autonomous cars, military operations, and robotics models should provide real-time explanations to ensure safety. However, the existing state-of-the-art XAIs have several challenges addressing this real-time explanation aspect. There are various factors for these XAI challenges. For example, a DNN may require numerous layers, thousands, millions, or billions of parameters to process the real-time input data, which is time-consuming, especially for large models. On the other hand, large volumes of data are generated continuously in real-time situations. For instance, autonomous cars have continuous and constant data streams from various sources, such as sensors, lidars, and radars with latency constraints. Hence, processing and providing a model explanation for this large amount of data and the latency constraint requires efficient XAI algorithms and techniques in real-time. Much effort is needed to address these challenges and meet the real-time requirements of safety-critical applications by considering model optimization, parallel processing, efficient XAI algorithms, hybrid approaches, and other handy techniques.

\subsection{Multilingual and Multicultural Explanation}
In recent years, large AI models that work on the diversity of languages and cultures have been employed. However, these AI models face several challenges because of the users' expectations, multilingual and multicultural nature. Hence, suitable XAI methods are crucial to providing meaningful model explanations for cultural variations, regional preferences, and language diversities that allow us to address harmful biases and keep sensitive cultural norms in diverse societal environments.

\section{Conclusion} \label{sectionConclusion}
The applications of complex AI models are widely integrated with our daily life activities in various aspects. Because of their complex nature, the demand for transparency, accountability, and trustworthiness is highly increased, specifically when these complex models make automated decisions that impact our lives differently. XAI techniques are required to explain why the model made that decision or prediction. In this survey, we explored the standard definitions and terminologies, the need for XAI, the beneficiaries of XAI, techniques of XAI, and applications of XAI in various fields based on the current state-of-the-art XAI works of literature. The study focuses on post-hoc model explainability. The taxonomy is designed to provide high-level insights for each XAI technique. We classified XAI methods into different categories based on different perspectives, such as training stage, scopes, and design methodologies. 
In the training stage, ante-hoc and post-hoc explanation techniques are two different ways to explain the inner workings of AI systems. 
Perturbation-based and gradient-based methods are two of the most common algorithmic design methodologies for the development of XAI techniques. 
We discussed different perturbation and gradient-based XAI methods based on their underlying mathematical principles and assumptions, as well as their applicability and limitations. Furthermore, we discuss the usage of XAI techniques to explain the decisions and predictions of ML models employed in natural language processing and computer vision application areas. Our objective is to provide a comprehensive review of the latest XAI techniques, insights, and application areas to XAI researchers, XAI practitioners, AI model designers and developers, and XAI beneficiaries who are interested in enhancing the trustworthiness, transparency, accountability, and fairness of their AI models. We also highlighted research gaps and challenges of XAI to strengthen the existing XAI methods and to give future research directions in the field.

\bibliographystyle{elsarticle-num}
\bibliography{bib.bib}
\end{document}